\pdfoutput=1

\documentclass[10pt,twocolumn,letterpaper]{article}
\usepackage[accsupp]{axessibility}

\usepackage[]{cvpr}      

\usepackage[utf8]{inputenc} 
\usepackage[T1]{fontenc}    
\usepackage{hyperref}       
\usepackage{url}            
\usepackage{booktabs}       
\usepackage{amsfonts}       
\usepackage{nicefrac}       
\usepackage{microtype}      
\usepackage{xcolor}         

\usepackage{amsmath}
\usepackage{amssymb}
\usepackage{mathtools}
\usepackage{amsthm}

\usepackage{graphicx}
\usepackage{amsmath}
\usepackage{amssymb}
\usepackage{kotex} 
\usepackage{adjustbox}
\usepackage{comment}
\usepackage{subfigure}
\usepackage{caption}
\usepackage{mathtools}
\usepackage{paralist} 
\usepackage{bm}
\usepackage{dsfont}
\usepackage{wrapfig, blindtext}
\usepackage{multirow}

\usepackage{amssymb}
\usepackage{pifont}

\usepackage{bbm}

\DeclarePairedDelimiter\floor{\lfloor}{\rfloor}

\newcommand\csm[1]{{\textcolor{black}{#1}}}

\theoremstyle{plain}

\theoremstyle{definition}

\theoremstyle{remark}

\newcommand{\cmark}{\ding{51}}%
\newcommand{\xmark}{\ding{55}}%

\usepackage[capitalize]{cleveref}
\crefname{section}{Sec.}{Secs.}
\Crefname{section}{Section}{Sections}
\Crefname{table}{Table}{Tables}
\crefname{table}{Tab.}{Tabs.}

\def\cvprPaperID{5208} 
\def\confName{CVPR}
\def\confYear{2023}

\begin{document}

\title{Rebalancing Batch Normalization for \\ Exemplar-based Class-Incremental Learning}

\author{
Sungmin Cha\textsuperscript{\rm 1},\  Sungjun Cho\textsuperscript{\rm 2},\  Dasol Hwang\textsuperscript{\rm 2},\  Sunwon Hong\textsuperscript{\rm 1}, \  Moontae Lee\textsuperscript{\rm 2,3},\  and Taesup Moon\textsuperscript{\rm 1,4,5}\thanks{Corresponding author (E-mail: \texttt{tsmoon@snu.ac.kr})}\vspace{.05in} \\
  \textsuperscript{\rm 1}Department of ECE, Seoul National University\ \
  \textsuperscript{\rm 2}LG AI Research \ \
  \textsuperscript{\rm 3}University of Illinois Chicago \\
  \textsuperscript{\rm 4}ASRI / INMC / IPAI / AIIS, Seoul National University\ \ 
  \textsuperscript{\rm 5}SNU-LG AI Research Center \\
  \texttt{sungmin.cha@snu.ac.kr},\ \ \texttt{\{sungjun.cho, dasol.hwang\}@lgresearch.ai}, \\ \texttt{zghdtnsz96@snu.ac.kr},\ \ \texttt{moontae.lee@lgresearch.ai}, \texttt{tsmoon@snu.ac.kr}
}

\maketitle

\begin{abstract}

Batch Normalization (BN) and its variants has been extensively studied for neural nets in various computer vision tasks, but relatively little work has been dedicated to studying the effect of BN in continual learning. To that end, we develop a new update patch for BN, particularly tailored for the exemplar-based class-incremental learning (CIL). The main issue of BN in CIL is the imbalance of training data between current and past tasks in a mini-batch, which makes the empirical mean and variance as well as the learnable affine transformation parameters of BN heavily biased toward the current task --- contributing to the forgetting of past tasks. While one of the recent BN variants has been developed for ``\textit{online}'' CIL, in which the training is done with a single epoch, we show that their method does not necessarily bring gains for ``offline'' CIL, in which a model is trained with multiple epochs on the imbalanced training data. The main reason for the ineffectiveness of their method lies in not fully addressing the data imbalance issue, especially in computing the gradients for learning the affine transformation parameters of BN. Accordingly, our new hyperparameter-free variant, dubbed as Task-Balanced BN (TBBN), is proposed to more correctly resolve the imbalance issue by making a horizontally-concatenated task-balanced batch using both reshape and repeat operations during training. Based on our experiments on class incremental learning of CIFAR-100, ImageNet-100, and five dissimilar task datasets, we demonstrate that our TBBN, which works exactly the same as the vanilla BN in the inference time, is easily applicable to most existing exemplar-based offline CIL algorithms and consistently outperforms other BN variants.


\end{abstract}

\vspace{-.2in}
\section{Introduction}

In recent years, continual learning (CL) has been actively studied to efficiently learn a neural network on sequentially arriving datasets while eliminating the process of re-training from scratch at each arrival of a new dataset~\cite{(cl_survey)delange2021continual}.
However, since the model is typically trained on a dataset that is heavily skewed toward the current task at each step, the resulting neural network often suffers from suboptimal trade-off between stability and plasticity~\cite{(tradeoff)mermillod2013stability} during CL. To overcome this issue, various studies have been focused on addressing the so-called \textit{catastrophic forgetting} phenomenon \cite{(cl_survey)delange2021continual, (cl_survey2)parisi2019continual}. 



Among different CL settings, the class-incremental learning (CIL) setting where the classifier needs to learn previously unseen classes at each incremental step has recently drawn attention due to its practicality \cite{(bic)wu2019large, (il2m)belouadah2019il2m, (lucir)hou2019learning, (ss-il)ahn2021ss, (podnet)douillard2020podnet, (eeil)castro2018end, (wa)zhao2020maintaining, (ccil)mittal2021essentials, (der)yan2021dynamically}. Most state-of-the-art CIL algorithms maintain a small \textit{exemplar memory} to store a subset of previously used training data and combine it with the current task dataset to mitigate the forgetting of past knowledge. A key issue of exemplar-based CIL is that the model prediction becomes heavily biased towards more recently learned classes, due to the imbalance between the training data from the current task (that are abundantly available) and past tasks (with limited access through exemplar memory). In response, recently proposed solutions to biased predictions include bias correction~\cite{(bic)wu2019large}, unified classifier~\cite{(lucir)hou2019learning}, and separated softmax~\cite{(ss-il)ahn2021ss}, which greatly improved the overall accuracy of CIL methods across all classes learned so far.

Despite such progress, relatively less focus has been made on backbone architectures under CIL on computer vision tasks. Specifically, the popular CNN-based models (\textit{e.g.}, ResNet~\cite{(resnet)he2016deep}) are mainly used as feature extractors, and those models are equipped with Batch Normalization (BN) ~\cite{(BN)ioffe2015batch} by default. However, since BN is designed for single-task training on CNNs, applying BN directly to exemplar-based CIL 
results in statistics biased toward the current task due to the imbalance between current and past tasks' data in a mini-batch. Recently, \cite{(CN)pham2022continual} pointed out this issue in CIL, dubbed as \textit{cross-task normalization effect}, and proposed a new normalization scheme called Continual Normalization (CN), which applies Group Normalization (GN) \cite{(GroupBN)wu2018group} across the channel dimension before running BN across the batch dimension. Therefore, the difference in feature distributions among tasks is essentially removed by GN, and the following BN computes task-balanced mean and variance statistics which is shown to outperform vanilla BN on \textit{online} CIL settings.

In this paper, we argue that CN only partially resolves the bias issue in exemplar-based CIL. Specifically, we find that the gradients on the affine transformation parameters remain biased towards the current task when using CN. As a result, it leads to inconsistent performance gains in the \textit{offline} CIL setting, which is considered more practical than \textit{online} CIL. 
To this end, we propose a simple yet novel hyperparameter-free normalization layer, dubbed as Task-Balanced Batch Normalization (TBBN), which effectively resolves the bias issue. Our method employs adaptive \textit{reshape} and \textit{repeat} operations on the mini-batch feature map during training in order to compute task-balanced normalization statistics and gradients for learning the affine transformation parameters. Our method does not require any hyperparameter as the size-inputs for reshape and repeat operations are determined adaptively. Furthermore, the application of TBBN during testing is identical to vanilla BN, requiring \textit{no} change on the backbone architecture. Through extensive offline CIL experiments on CIFAR-100, ImageNet-100, and five dissimilar task datasets, we show that a simple replacement of BN layers in the backbone CNN model with TBBN benefits most state-of-the-art exemplar-based CIL algorithms towards an additional boost in performance. Our analysis shows that the gain of TBBN is consistent across various backbone architectures and datasets, suggesting its potential to become a correct choice for exemplar-based offline CIL algorithms. 

\section{Related Work}
\noindent\textbf{Exemplar-based CIL.} \ \
Among various work in offline CIL, iCaRL~\cite{(icarl)rebuffi2017icarl} is the first to propose an exemplar-based method via utilization of nearest-mean-of-exemplars classification and representation learning.
EEIL~\cite{(eeil)castro2018end} next devised a distillation-based CIL method with both balanced fine-tuning and representative memory updating.
BiC~\cite{(bic)wu2019large} focused on the biased prediction aspect due to data imbalance in CIL and proposed a novel yet simple idea to attach an additional layer that corrects biased predictions. 
Inspired by LWF~\cite{(lwf)li2017learning}, LUCIR~\cite{(lucir)hou2019learning} developed a method that overcomes catastrophic forgetting via a more sophisticated cosine normalization-based loss function. 
PODNet~\cite{(podnet)douillard2020podnet} proposed another distillation-based method that focuses on spatial-based loss to reduce representation forgetting. 
Most recently, two state-of-the-art knowledge distillation-based CIL methods have been proposed. First, SS-IL~\cite{(ss-il)ahn2021ss} demonstrated that a pure softmax mainly causes biased predictions in CIL, leading to the Separated Softmax layer that achieved significant performance gains. Second, AFC~\cite{(AFC)kang2022class} devised a novel regularization for knowledge distillation to minimize the changes of important features when learning new tasks.

On the other hand, online CIL also has attracted attention~\cite{shim2021online,(OCL)https://doi.org/10.48550/arxiv.1908.04742}, but their performance is significantly lower in the benchmark (\textit{e.g.}, CIFAR-100) than offline CIL counterparts, as reported in the survey paper~\cite{(online_cl_survey)mai2021online}.

\noindent\textbf{Normalization layer.} \ \ 
\csm{One} initial motivation \csm{for} devising Batch Normalization (BN) \cite{(BN)ioffe2015batch} was to address internal covariate shift of a neural net. 
However, the belief in internal covariate shift was broken by follow-up studies, and the benefits from BN in terms of training perspective were analyzed in various directions~\cite{(bn_analysis1)bjorck2018understanding, (bn_analysis2)santurkar2018does}.
\csm{Thereafter}, several normalization layers devised for various computer vision tasks have been proposed, \csm{each} with their respective advantage\csm{s} ~\cite{(IN)ulyanov2016instance, (GroupBN)wu2018group, (GhostBN)hoffer2017train, (SN)luo2019switchable, (BRN)ioffe2017batch}, and for more diverse tasks, such as meta-learning~\cite{(tasknorm)bronskill2020tasknorm}, domain adaptation~\cite{(transbn)wang2019transferable}, task-incremental learning~\cite{(janghyeon_BN)lee2020continual}, and online CL~\cite{(CN)pham2022continual}.

\section{Preliminaries}

\noindent\textbf{Notation.} \ \ We assume the setting of offline CIL as proposed in \cite{(ss-il)ahn2021ss}; 
namely, a dataset $\mathcal{D}_t$ for each task $t$ arrives incrementally, and we use $T$ to denote the total number of tasks. 
Each $\mathcal{D}_t$ consists of pairs of an input image $\bm{x}_t$ and its target label $y_t$.
We assume that each incrementally added task introduces $m$ new classes that the model has never observed previously, and hence the total number of classes observed until task $t$ equals $C_t=m\cdot t$.
Therefore, the target label in each task is labeled as $y_t\in\{{C}_{t-1}+1, \dots, {C}_t\}\triangleq\mathcal{C}_t$.
To save exemplars from previous tasks, we use exemplar memory denoted by $\mathcal{M}$.
While $\mathcal{M}$ is updated via a sampling algorithm after training on each task, the memory size is set to not exceed $|\mathcal{M}|$ over the entire course of tasks. 
In particular, exemplar memory $\mathcal{M}_{t-1}$ maintains $\floor*{\frac{|\mathcal{M}|}{C_{t-1}}}$ images per each past class and is used for training task $t$.
To merge instances from current as well as previous tasks, we consider a training mini-batch size of $B = B_c + B_p$, where $B_c$ and $B_p$ denote the number of data samples from $\mathcal{D}_t$ and $\mathcal{M}_{t-1}$, respectively. 
Without loss of generality, we assume that each mini-batch $(\bm{x}^{{B}}, y^{{B}}) \sim (\mathcal{D}_t \cup \mathcal{M}_{t-1})$ is a concatenation of $B_c$ samples from $\mathcal{D}_t$ with $B_p$ samples from $\mathcal{M}_{t-1}$: $\bm{x}^{{B}} = [\bm{x}_1; \dots; \bm{x}_{B_c}; \bm{x}_{B_c + 1}; \dots; \bm{x}_{B}]$ and $y^{{B}} = [y_1; \dots; y_{B_c}; y_{B_c + 1}; \dots; y_{B}]$ with $\{(\bm{x}_i,y_i)\}_{i=1}^{B_c} \subset \mathcal{D}_t$ and $\{(\bm{x}_i,y_i)\}_{i=B_c+1}^{B} \subset \mathcal{M}_{t-1}$.
Note that $B_c$ is greater than $B_p$ in typical CIL settings, and it is this imbalance that is generally known to cause predictions biased towards the current task \cite{(ss-il)ahn2021ss, (il2m)belouadah2019il2m, (bic)wu2019large}.
Therefore, we set the sampling ratio between $B_c$ and $B_p$ as $B_c : B_p = 3 : 1$ in our experiments.
Lastly, we use $\bm{h}\in\mathbb{R}^{B \times C \times D}$ to denote the input feature map for an intermediate BN layer given $\bm{x}^{{B}}$ as input. Here, $C$ and $D$ denote the number of channels and feature dimension, respectively.
\noindent\textbf{Batch Normalization.} \ \ Due to its powerful practicality and efficiency, Batch Normalization (BN)~\cite{(BN)ioffe2015batch} has been the go-to normalization layer across various state-of-the-art neural network architectures \cite{(resnet)he2016deep, (inceptionnet)szegedy2016rethinking, (efficientnet)tan2019efficientnet}.
During training, BN calculates the empirical mean $\hat{\mu} \in \mathbb{R}^C$ and variance $\hat{\sigma}^2 \in \mathbb{R}^{C}$ of the given mini-batch as
\vspace{-.05in}
\begin{equation}
\hat{\mu}_{c}=\frac{1}{B D} \sum^{B}_{b=1} \sum^{D}_{d=1} \bm{h}_{b,c,d},
\ \hat{\sigma}^2_{c}=\frac{1}{B D} \sum^{B}_{b=1} \sum^{D}_{d=1} (\bm{h}_{b,c,d} - \hat{\mu}_{c})^2.\label{eqn:mean_var} 
\end{equation}
The feature map $\bm{h}$ is then normalized to have zero-mean unit-variance using $\hat{\mu}$ and $\hat{\sigma}^2$ to obtain $\hat{\bm{h}}\in\mathbb{R}^{B \times C \times D}$. The normalized feature $\hat{\bm{h}}$ is lastly affine-transformed with a trainable scale and shift parameters $\gamma, \beta \in \mathbb{R}^{C}$ to $\bm{y}$:
\begin{gather}
    \hat{\bm{h}}_{b}= \frac{\bm{h}_{b} - \hat{\mu}\cdot \mathds{1}^\intercal}{\sqrt{\hat{\sigma}^2\cdot \mathds{1}^\intercal + \epsilon\cdot \mathds{1}^\intercal}}
    \label{eqn:normalization}\\
    \bm{y}_{b}= \gamma\cdot \mathds{1}^\intercal \odot \hat{\bm{h}}_{b} + \beta\cdot \mathds{1}^\intercal, \label{eqn:affine_transformation}
\end{gather}
in which $\mathds{1} \in \mathbb{R}^{D}$ is the all-1 vector, $\odot$ denotes the element-wise multiplication, and $b$ is the data index within the mini-batch. Note the division and $\sqrt{\cdot}$ in (\ref{eqn:normalization}) are also done in an element-wise sense. The $\epsilon$ term is a small value added for numerical stability, which we set as $\epsilon = 10^{-5}$ in the experiments. 
It is worth noting that $\gamma$ and $\beta$ are trained by computing the gradients 
\begin{equation}
\frac{\partial \mathcal{L}}{\partial \gamma_{c}} = \sum^{B,D}_{b,d=1} \frac{\partial \mathcal{L}}{\partial \bm{y}_{b,c,d}} \odot \hat{\bm{h}}_{b,c,d}, \ \ \
\frac{\partial \mathcal{L}}{\partial \beta_{c}} = \sum^{B,D}_{b,d=1} \frac{\partial \mathcal{L}}{\partial \bm{y}_{b,c,d}}\label{eqn:gradient_gamma_beta}
\end{equation}
under a loss function $\mathcal{L}$.
For the normalization in the test phase, running $\mu$ and $\sigma^{2}$ are updated incrementally at each $i$-th training iteration via exponential moving average:
\vspace{-.05in}
\begin{equation}
\begin{aligned}
& \mu_{(i)} = \alpha \cdot \mathds{1}^\intercal \odot \hat{\mu}_{(i)} +  (1-\alpha) \cdot \mathds{1}^\intercal \odot \mu_{(i-1)} \label{eqn:moving_average}\\ 
& \sigma^{2}_{(i)} = \alpha \cdot \mathds{1}^\intercal \odot \hat{\sigma}^2_{(i)} +  \left((1-\alpha) \dfrac{V - 1}{V}
\right) \cdot \mathds{1}^\intercal \odot   \sigma^2_{(i-1)}
\end{aligned}
\end{equation}
in which $V = B\cdot D$ represents the vessel's correction, and $\alpha$ is the momentum hyperparameter generally set as $0.1$.


















\section{Main Method}
\subsection{Motivation}
\vspace{-.05in}
While exemplar memory is essential for achieving state-of-the-art performance in offline CIL~\cite{(cil_survey)masana2020class}, most methods use well-known backbone classification models (\textit{e.g.}, ResNet-18~\cite{(resnet)he2016deep}) with vanilla BN layers without much scrutiny over the issue of data-imbalance between current and previous tasks. One notable exception is Continual Normalization (CN)~\cite{(CN)pham2022continual}, but CN has only shown promise in online CIL settings where the data-imbalance causes relatively less of an impact than in offline CIL due to the \textit{single} epoch learning. It is not clear whether CN is also effective in offline CIL, where the imbalanced mini-batches are used for \textit{multiple} epochs.  
We conjecture that while CN performs task-balancing by normalizing each channel (or groups of channels) along the spatial dimension, this comes at a cost in discriminative power as it normalizes features from all tasks equally without any separation between tasks. In a later section, we find through experimentation that this limitation is especially detrimental in offline CIL, as CN leads to negligible gain over BN or even worse performance in some cases.

\begin{figure*}[t]
\centering 
\subfigure[True population distribution]
{\includegraphics[width=0.23\linewidth]{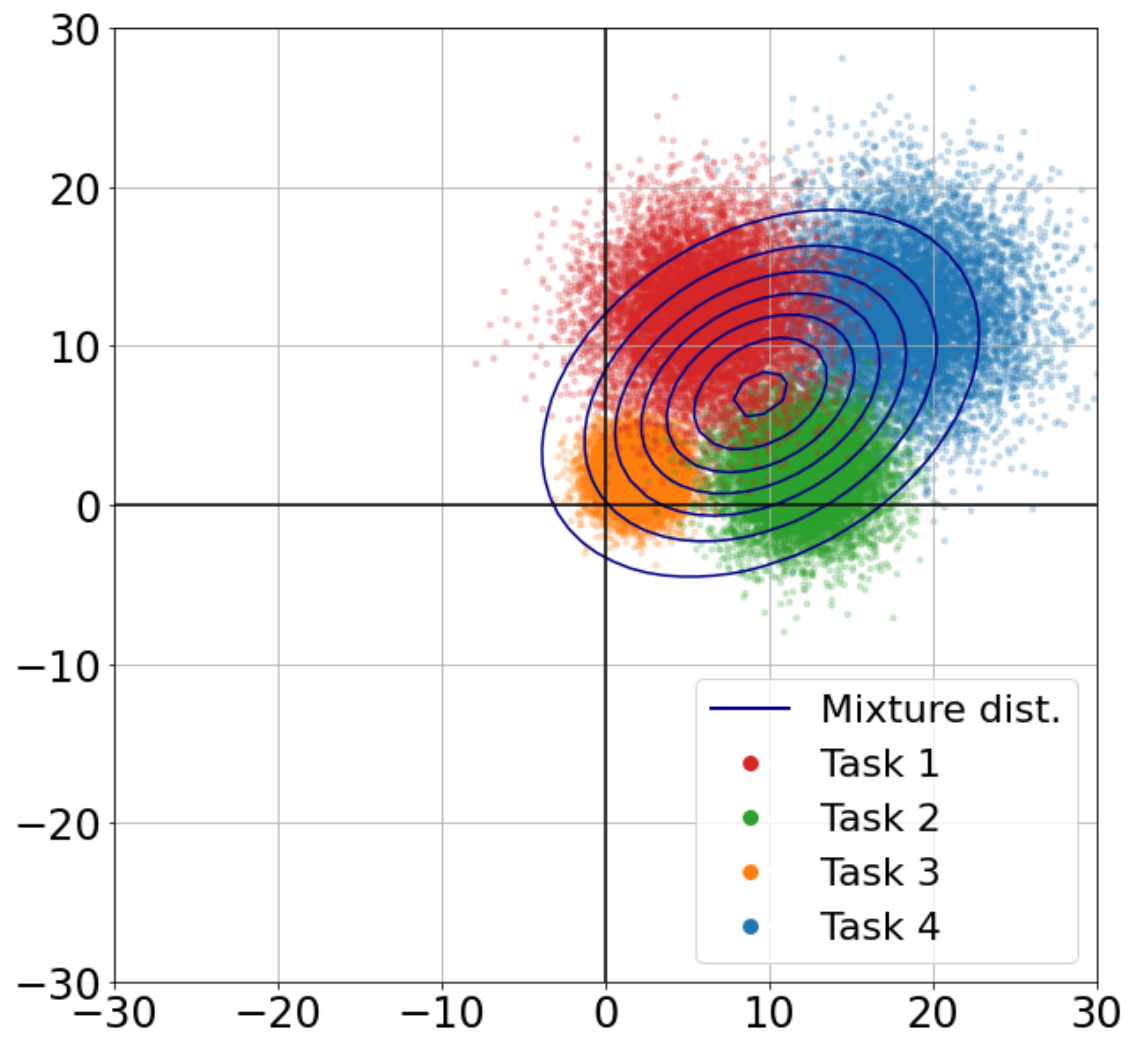}
\label{figure:illustration1}}
\subfigure[Biased train data during CIL]
{\includegraphics[width=0.23\linewidth]{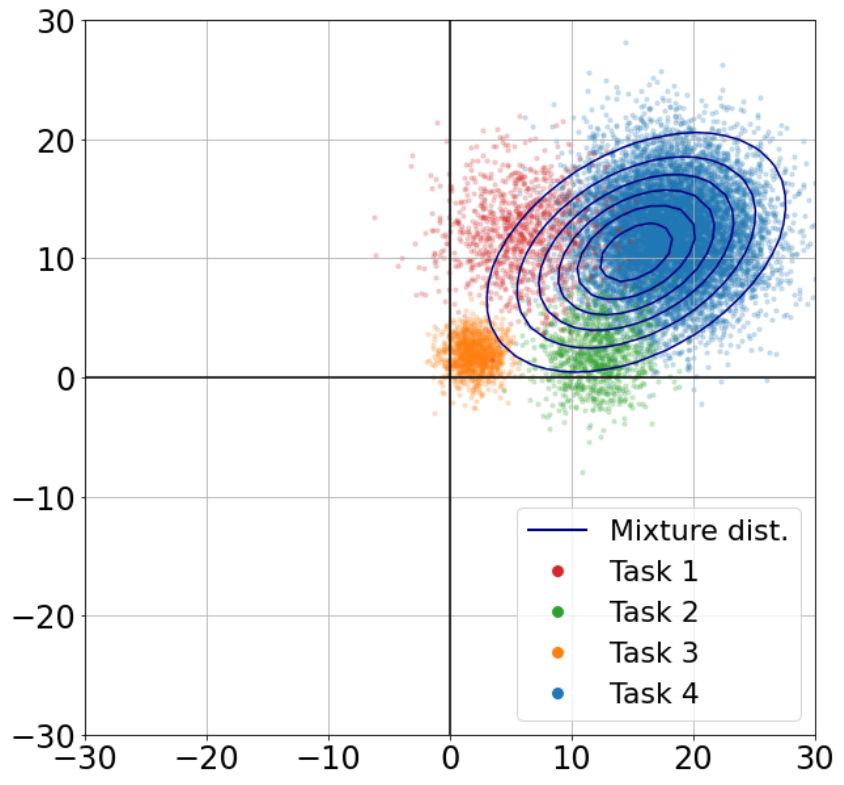}
\label{figure:illustration2}}
\subfigure[Test data normalized via CN]
{\includegraphics[width=0.22\linewidth]{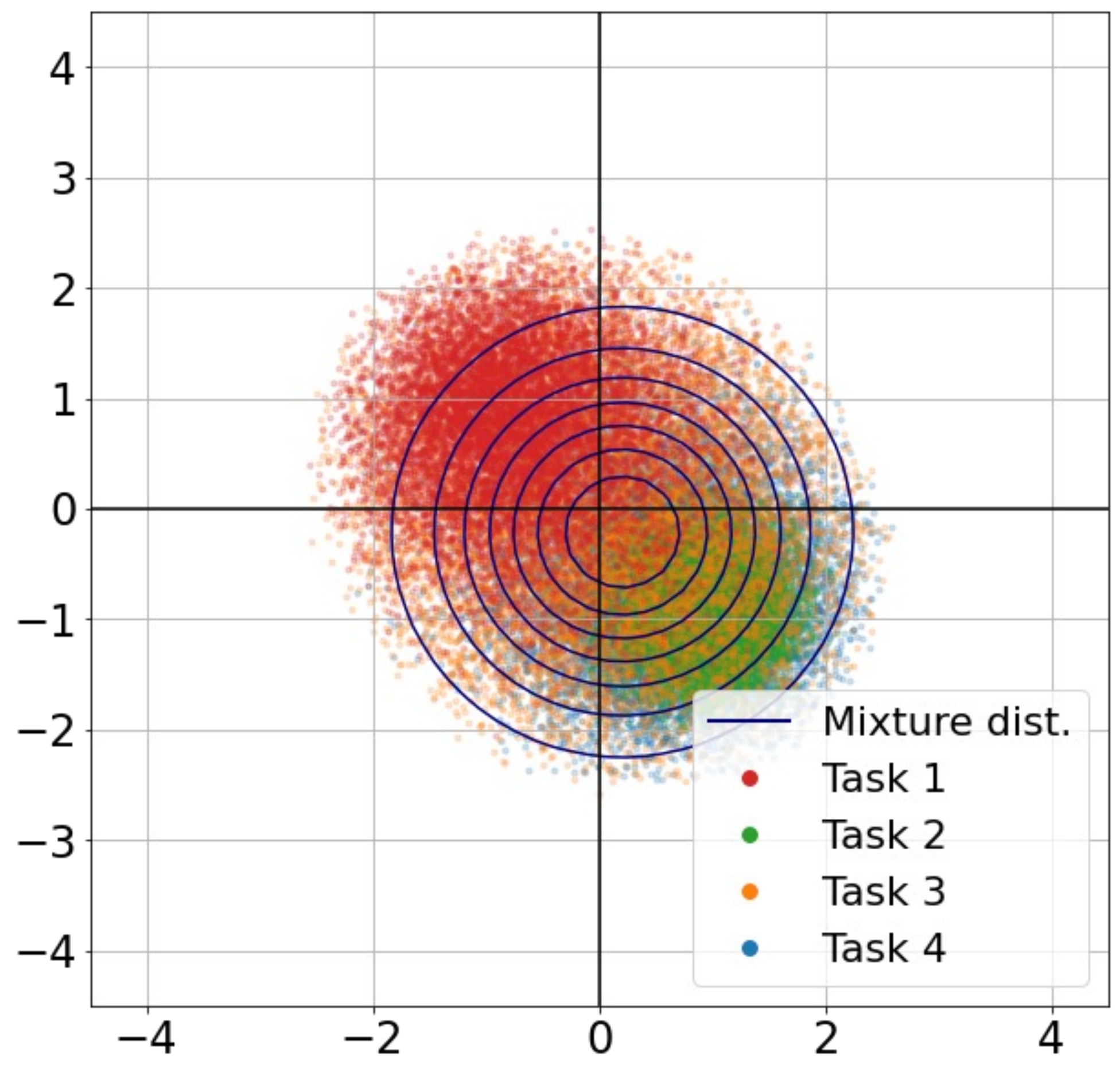}
\label{figure:illustration3}}
\subfigure[Test data normalized via TBBN]
{\includegraphics[width=0.22\linewidth]{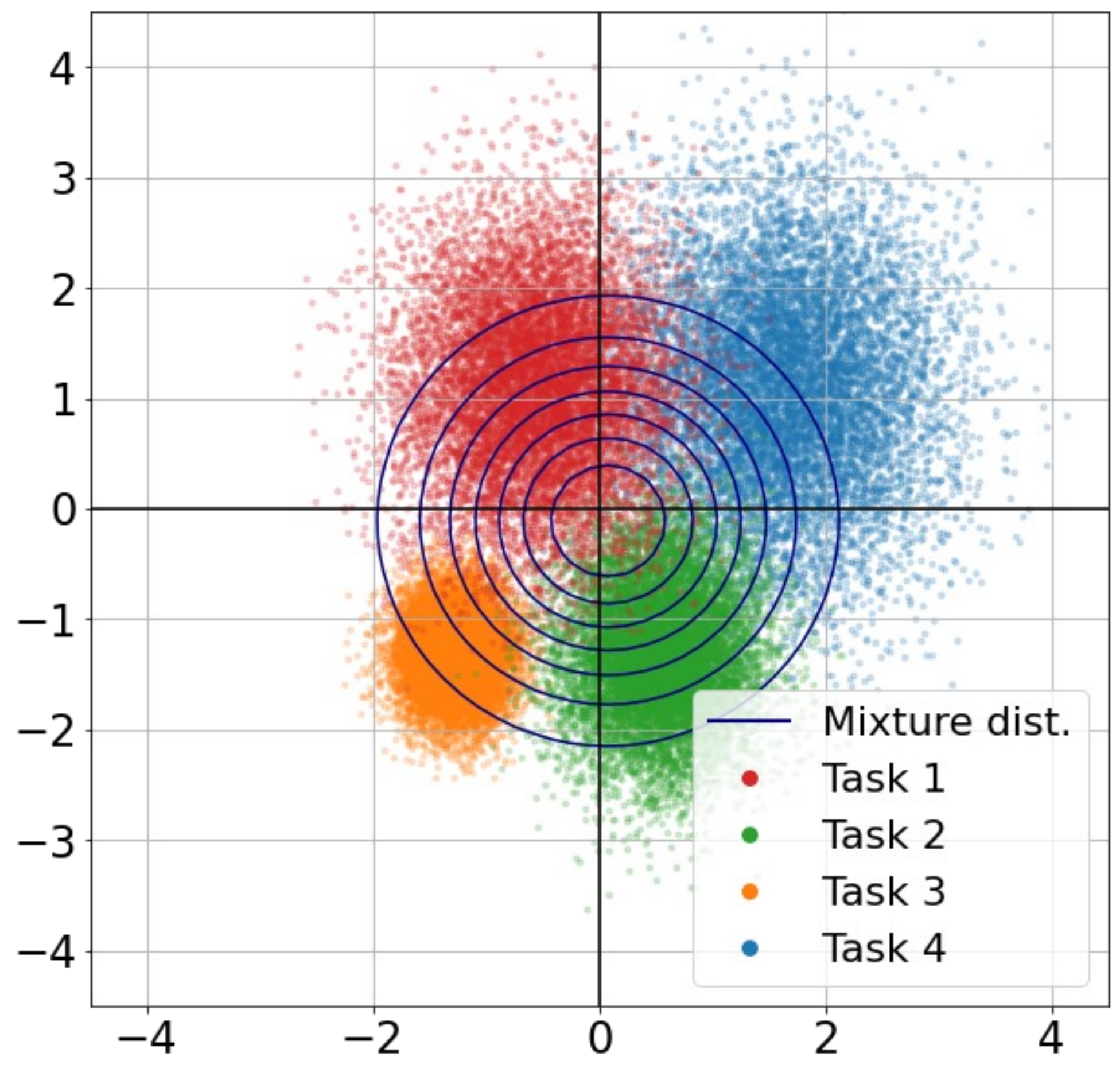}
\label{figure:illustration4}}
\vspace{-.15in}
\caption{(a) 2-D visualizations on a mixture of four Gaussian distributions, each representing a feature map distribution of each task. (b) Imbalanced training data in exemplar-based CIL while learning Task 4, deviating from true test distribution. (c) Test set normalization result of CN that computes normalization statistics from data in (b). (d) Test set normalization result of our proposed TBBN that computes normalization statistics from data in (b).}
\vspace{-.2in}
\label{figure:illustration}
 \end{figure*}


For a concrete motivation, we consider a toy example to reflect upon the necessity of devising a correct BN mechanism for the exemplar memory-based CIL.
Consider a synthetic input feature distribution for an intermediate BN layer shown in Figure \ref{figure:illustration}.
Figure \ref{figure:illustration1} visualizes 2-D data samples generated from a mixture of four different Gaussian distributions, each representing the feature distribution for each given task.\footnote{In fact, we generated 20 dimensional Gaussian vectors, which is generated by concatenating i.i.d. 2-D Gaussians 10 times, and only the first two dimensions are shown in the figures. The normalization by CN and TBBN shown in Figure 1(c)(d) are done by obtaining the statistics from the entire 20 dimensions (which exemplifies the channel dimension in a feature map). }
In a joint training setting, the mini-batch samples are uniformly sampled across four tasks, and thus the empirical mean and variance obtained from BN would approach the true population mean and variance. Such mean and variance will then successfully standardize the samples in a mini-batch during both training and testing phases via the procedure given in (\ref{eqn:normalization}) and (\ref{eqn:affine_transformation}). 

In an exemplar memory-based CIL setting, however, there is a severe imbalance among samples in the training mini-batch, since the majority of the samples comes from the current task (Task 4) as shown in Figure \ref{figure:illustration2}. In this case, the mean and variance calculated by the BN layer is clearly biased towards those of the current task, failing to match the true statistics at test time.
In other words, using the biased mean and variance obtained during training to normalize test samples leads the samples to deviate from a standardized distribution.
It is evident that such mismatch between the train and test sample distributions can cause performance deterioration since the trainable parameters in BN are obtained from normalized \textit{training} samples. 
An obvious remedy would be to obtain the \textit{task-balanced} empirical mean and variance during training, and use them in the BN layer to normalize the test samples. Figure \ref{figure:illustration3} and \ref{figure:illustration4} highlight the difference between the test normalization done by CN and our TBBN, respectively, which computes the empirical mean and variance from the mini-batch in Figure \ref{figure:illustration2}. Note that while both successfully normalize test samples in a task-balanced way, CN tends to lose the discriminative structure among the tasks due to the intermediate GN step. In constrast, our TBBN, as will be described in details below, normalizes the test data features while maintaining the discriminative structure.

In addition to the mismatch during test time, a na\"ive application of BN on the biased mini-batch samples can also cause an issue during training. That is, when training for a current task (\textit{e.g.}, Task 4 in Figure \ref{figure:illustration}) is finished, the empirical mean and variance obtained by a BN layer would be heavily biased towards the most recent task. When a new task arrives (say, Task 5), since only a small number of samples from Task 4 is saved in the exemplar memory, the new empirical mean and variance will now become biased towards Task 5, drifting away from the previous statistics updated via the exponential moving average in (\ref{eqn:moving_average}). This mismatch during training can also significantly alter the learned representations of the past task samples in the exemplar memory after normalization. Therefore, this necessitates re-training of subsequent layers, unintentionally causing forgetting of previously learned representations. 



\begin{figure}[h]
\vspace{-.1in}
\centering
\includegraphics[width=0.8\linewidth]{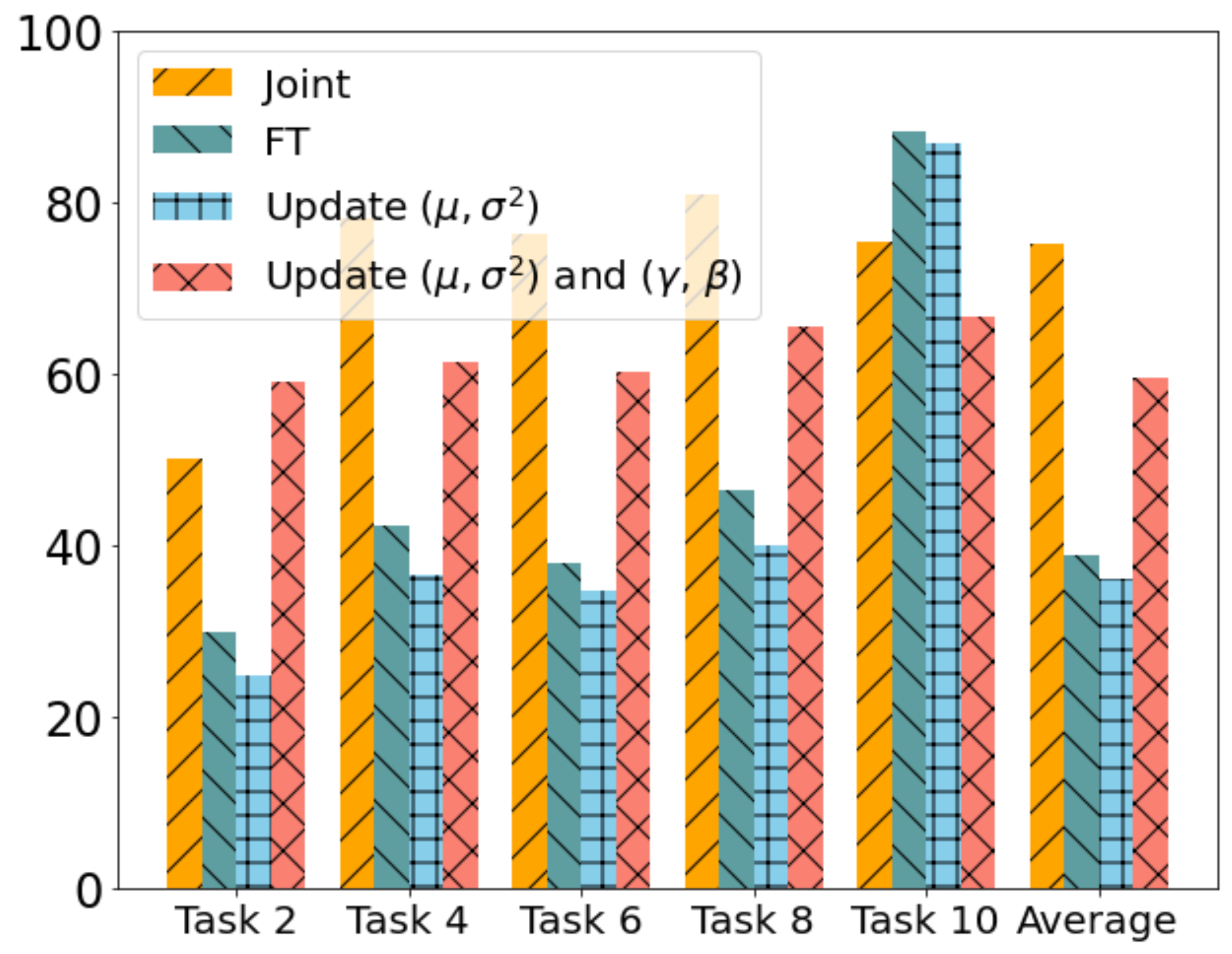}
\vspace{-.1in}
\caption{Individual task and average accuracies on ImageNet-100 split into 10 tasks for CIL. $y$-axis shows the accuracy for each task and the average accuracy across all tasks after full training. Recomputing the affine-transformation parameters $(\gamma, \beta)$ as well as empirical statistics $(\mu, \sigma^2)$ based on the full dataset leads to significant boost in performance.}
\vspace{-.1in}
\label{figure:motivation_experiments}
 \end{figure}

With regards to the two aforementioned issues, we argue that it is crucial to calculate the \textit{task-balanced} mean and variance for BN layers in exemplar \csm{memory-based} CIL.
While a similar observation has also been made in the recent work \cite{(CN)pham2022continual}, we go one step further and note that the affine transformation parameters of BN (\textit{i.e.} $\gamma$ and $\beta$ in (\ref{eqn:affine_transformation})) should also be learned in a task-balanced manner. 
For a demonstration purpose, we train a ResNet-18 model with vanilla BN and exemplar memory of size $|\mathcal{M}| = 2000$ on ImageNet-100 split to 10 tasks with 10 classes per task.
We test four different CIL procedures including two oracle-based approaches that can access the entire training set during test time to exhibit the necessity of task-balanced statistics and affine-transformations.
In Figure \ref{figure:motivation_experiments}, \textsc{Joint} denotes the algorithm that jointly trains the model on all \csm{tasks} at once, providing an upper bound in average accuracy. On the other hand, \textsc{FT} is the fine-tuning baseline that simply fine-tunes the model on samples from the current task and exemplar memory for each task. Note that \textsc{FT} performs poorly on previous tasks except for the final task, clearly reflecting the issue of biased predictions. \textsc{Update $(\mu,\sigma^2)$} is an oracle scheme that \textit{freezes} all trainable parameters of the trained \textsc{FT} model, and only recomputes the mean and variance of all BN layers using the entire dataset such that they match the task-balanced population statistics. Finally, \textsc{Update $(\mu,\sigma^2)$ and $(\gamma,\beta)$} denotes another oracle scheme that also \textit{re-trains} the $(\gamma,\beta)$ parameters for all BN layers on the entire dataset, in addition to recomputing the $(\mu,\sigma^2)$ statistics. 

Figure \ref{figure:motivation_experiments} displays the classification accuracy of four methods on every 2nd task and the average accuracy across all tasks, all measured after training up to the final task. 
The slight drop in performance of \textsc{Update $(\mu, \sigma^2)$} compared to \textsc{FT} shows that only obtaining the task-balanced $(\mu,\sigma^2)$ is not sufficient due to mismatch between the updated $(\mu,\sigma^2)$ and the frozen $(\gamma,\beta)$. In contrast, when affine-transformation parameters are also recomputed with task-balanced data, we observe a significant boost in average accuracy. This is particularly interesting as the affine-transformation parameters account for less than 1\% of the entire model parameters, and simply making them task-balanced along with the empirical $(\mu,\sigma^2)$ statistics in BN layers can bring a significant performance boost under a simple FT baseline. Inspired by this example, we develop our Task-Balanced Batch Normalization (TBBN) layer in the next section.

\begin{figure*}[h]
\centering 
\includegraphics[width=0.9\linewidth]{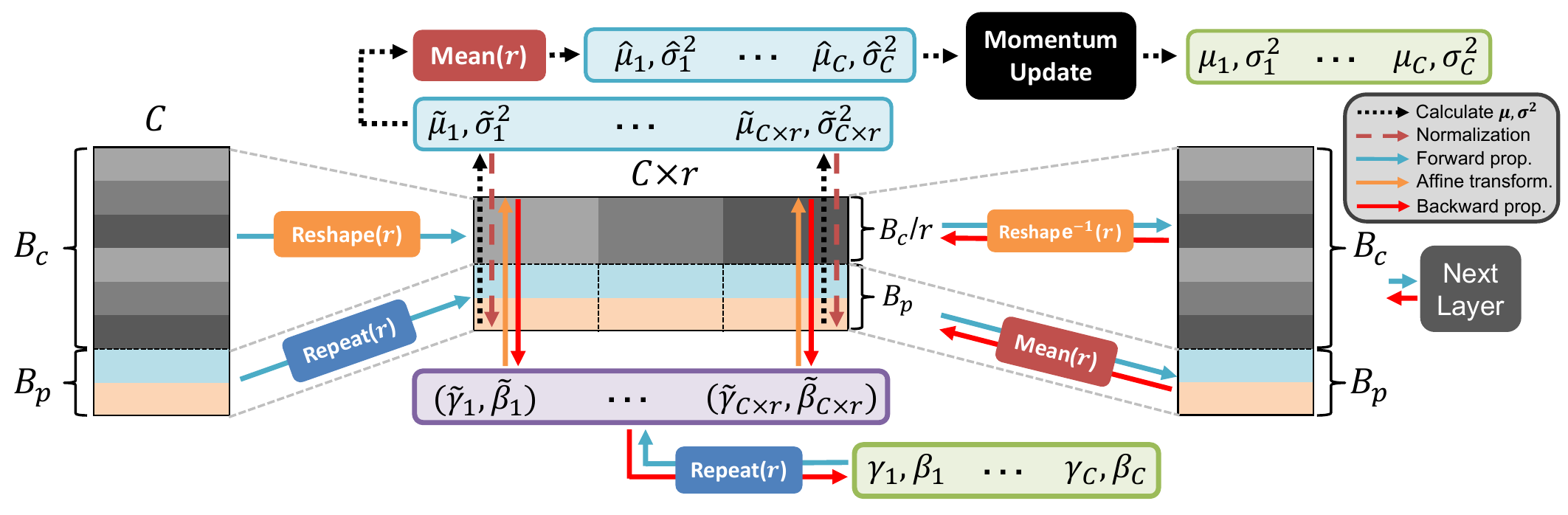}
\vspace{-.15in}
\caption{Illustration of the forward (blue arrows) and backward (red arrows) propagations of TBBN. By applying both reshape and repeat operations to a given input of BN layer, TBBN induces that task-balanced calculation and training of  $(\mu, \sigma^2)$ and $(\gamma, \beta)$, respectively.  
}
\label{figure:tbbn_figure}
\vspace{-.2in}
 \end{figure*}

\subsection{Task-Balanced Batch Normalization (TBBN)}
\vspace{-.05in}
To solve the problem of BN in the exemplar-based CIL, we propose Task-Balanced Batch Normalization (TBBN) consisting of two components: 1) Task-balanced $\mu$ and $\sigma^2$ calculation and 2) Task-balanced training of $\gamma$ and $\beta$.

\noindent\textbf{Task-balanced $\mu$ and $\sigma^2$ calculation.}  \ \
As discussed in the previous section, the class imbalance between the current and previous data biases \csm{the} mean and variance toward the current task.
Instead, we propose a task-balanced algorithm that better computes the empirical mean and variance in the current batch. 
As shown in Figure \ref{figure:tbbn_figure}, we design the two different tensor operations: \textit{reshape} and \textit{repeat}, applying them to the current and previous parts, respectively.
We determine the number of repeat/reshape operations $r$ by the following \eqref{eqn:task_adaptive_r}. 
It determines the value of $r$, which is used to balance the number of batches sampled from the exemplar memory ($\frac{B_{p}}{t-1}$) and the number of batches of the current task ($\frac{B_{c}}{r}$) after reshaping with $r$, so that the ratio between them becomes 1:1. 
Note that $r$ is a hyperparameter that need not be externally tuned, but \csm{that} our method provides a task-adaptive guidance for automatically selecting $r$:
\begin{equation}
    \frac{B_{c}}{r} = \frac{B_{p}}{t-1}, \ \text{hence,} \ \ r = \frac{B_{c}}{B_{p}} \cdot (t-1),\label{eqn:task_adaptive_r}
\end{equation}
in which $t = 2, \dots, T$, and $r$ is set to 1 when $t = 1$.


Let $\bm{h}_{B_c}$ be the input feature map of a BN layer for the current task $t$.
After applying the reshape operation ${\bm{{\mathrm{H}}}}_{B_c} = F_{S}(\bm{h}_{B_c} ; r):\mathbb{R}^{B_c \times C \times D} \rightarrow \mathbb{R}^{B_{c}/r \times C\cdot r \times D}$ with $r$, we divide $\bm{h}_{B_c}$ into $r$ splits, where each split has $\frac{B_{c}}{r}$ numbers of data from the current task.
Similarly, let $\bm{h}_{B_p}$ denote the data from the previous tasks.
As $\bm{h}_{B_p}$ consists of the data uniformly sampled from task 1 to $t-1$, $\frac{B_{p}}{t-1}$ numbers of data belongs to $\bm{h}_{B_p}$ for each previous task in expectation.
The key insight is that we can make every split balanced by repeating $\bm{h}_{B_p}$ by $r$ times if the task ratios $\frac{B_{c}}{r}$ and $\frac{B_{p}}{t-1}$ become equivalent.
So that, we repeat $\bm{h}_{B_p}$ for $r$ times by applying a tensor repeat function ${\bm{{\mathrm{H}}}}_{B_p}=F_{P}(\bm{h}_{B_p} ; r):\mathbb{R}^{B_p \times C \times D} \rightarrow \mathbb{R}^{B_p \times C\cdot r \times D}$.
Then, we concatenate both tensors along batch axis, $\bm{\mathrm{H}} = ( {\bm{{\mathrm{H}}}}_{B_c} ; {\bm{{\mathrm{H}}}}_{B_p} )\in\mathbb{R}^{(B_c/r + B_p) \times C\cdot r \times D}$, and calculate a task-balanced empirical mean ${\tilde{\mu}}\in\mathbb{R}^{C\cdot r}$ and variance ${\tilde{\sigma}}^2\in\mathbb{R}^{C\cdot r}$ from a horizontally-concatenated task-balanced batch $\bm{{\mathrm{H}}}$.
To update running mean ${{\mu}}\in\mathbb{R}^{C}$ and variance ${{\sigma}}^2\in\mathbb{R}^{C}$ for test phase, we average ${\tilde{\mu}}$ and ${\tilde{\sigma}}^2$ over $r$ splits, later applying these to \csm{the }exponential moving average (\ref{eqn:moving_average}) \csm{to update}  ${{\mu}}$ and ${{\sigma}^2}$.

\noindent\textbf{Task-balanced training of $\gamma$ and $\beta$.} \ \
Additionally, we propose a way to train learnable parameters of BN ($\gamma$ and $\beta$) in a task-balanced way to the current task.
After calculating the task-balanced ${\tilde{\mu}}$ and ${\tilde{\sigma}}^2$ from $\bm{{\mathrm{H}}}$, we apply these to normalize the horizontally-concatenated task-balanced batch $\bm{{\mathrm{H}}}$ via (\ref{eqn:normalization}).
Then, the normalized feature map $\tilde{\bm{{\mathrm{H}}}}\in\mathbb{R}^{(B_c/r + B_p) \times C\cdot r \times D}$ is affine-transformed by (\ref{eqn:affine_transformation}) with $\tilde{\gamma} = F_{RP}(\gamma)\in\mathbb{R}^{C\cdot r}$ and $\tilde{\beta} = F_{RP}(\beta)\in\mathbb{R}^{C\cdot r}$, resulting in the affine-tranformed feature ${\bm{{\mathrm{Y}}}} = ( {\bm{{\mathrm{Y}}}}_{B_c} ; {\bm{{\mathrm{Y}}}}_{B_p} )\in\mathbb{R}^{(B_c/r + B_p) \times C\cdot r \times D}$.
However, due to the different size of batch axis with the original batch size ($B_c/r + B_p \neq B$), ${\bm{{\mathrm{Y}}}}$ cannot be directly forward-propagated toward a next layer.
To solve this problem, first, we apply a reverse operation of reshape to  $\bm{\mathrm{Y}}_{B_c}$ to obtain $\bm{y}_{B_c} \in \mathbb{R}^{ B_c \times C \times D}$, where $F_{S}^{-1}(\bm{\mathrm{Y}}_{B_c}; r) :\mathbb{R}^{B/r \times C\cdot r \times D} \rightarrow\mathbb{R}^{ B_c \times C \times D}$.
Next, for the case of ${\bm{{\mathrm{Y}}}}_{B_p}$, we get $\bm{y}_{B_p} \in \mathbb{R}^{ B_p \times C \times D}$ by averaging ${\bm{{\mathrm{Y}}}}_{B_p}$  over $r$ repeats along the axis of channel, such as $\bm{y}_{B_p} = F_{A}(\bm{\mathrm{Y}}_{B_p}; r) :\mathbb{R}^{B_p \times C\cdot r \times D} \rightarrow\mathbb{R}^{ B_p \times C \times D}$. Finally, we forward-propagate the concatenated feature $\bm{y} = (\bm{{y}}_{B_c};\bm{{y}}_{B_p}) \in \mathbb{R}^{ B \times C \times D}$ toward the next layer.
\vspace{-.1in}

\subsubsection{Why does TBBN work?}
\vspace{-.05in}
\noindent\textbf{Calculating task-balanced $\mu$ and $\sigma^2$.} \ \
When performing the standard BN while fixing the number of data instances from previous tasks as $B_p$, the individual statistics $(\mu^{BN}, \sigma^{BN})$ becomes dominated by the current task. As the exponential moving average iteratively averages over biased empirical statistics, BN fails to asymptotically recover population mean and variance. This is especially clear when expanding the BN mean $\mu_t^{BN}$ during training on task $t$ in terms of means $\mu_i$ of the $i$-th task. Assuming that each task contributes uniformly in the global data distribution, $\mu_t^{BN}$ deviates from the true global mean $\mu_t^*$ by  
\begin{gather}
    \mu_t^* - \mu_t^{BN} = \dfrac{B - B_c t}{t(t-1)B}\sum_{i=1}^t \mu_i + \dfrac{B_c t - B}{(t-1)B} \mu_t.
\end{gather}
In a nutshell, BN fails to recover the true population mean unless the mini-batch is balanced such that $B_c = B/t$. Our proposed TBBN creates $r$ task-balanced bags of bootstrapped samples such that this balancing is satified, and approximates the population statistics $(\mu^*, \sigma^{2*})$ in an unbiased manner.






\noindent\textbf{Training $\gamma$ and $\beta$ with task-balanced gradients.} \ \
Note that the gradient for training $\beta$ and $\gamma$ of the original BN, shown in (\ref{eqn:gradient_gamma_beta}), is biased toward the current task due to the imbalance between $B_c$ and $B_p$ in the mini-batch.
On the other hand, the above task-balanced process makes both $\beta$ and $\gamma$ to be trained in a task-balanced way during backpropagation.
First, a gradient of $\bm{y}$ is back-propagated toward the BN layer and it can be denoted as: $\frac{\partial \mathcal{L}}{\partial \bm{y}} = (\frac{\partial \mathcal{L}}{\partial \bm{\mathrm{Y}}_{B_c}};\frac{\partial \mathcal{L}}{\partial \bm{\mathrm{Y}}_{B_p}}) \in \mathbb{R}^{ B \times C \times D}$.
    Then, the gradients of each current and task's batch pass the backward operation of $F_{S}^{-1}$ and $F_{A}$, respectively. As a result, the gradient of $\bm{\mathrm{Y}}$ becomes:
\begin{equation}
\vspace{-.1in}
\frac{\partial \mathcal{L}}{\partial \bm{\mathrm{Y}}} = \Big(F_{S,g}^{-1}\big(\frac{\partial \mathcal{L}}{\partial \bm{\mathrm{Y}}_{B_c}}; r\big);F_{A,g}\big(\frac{\partial \mathcal{L}}{\partial \bm{\mathrm{Y}}_{B_p}}; r\big)\Big),\label{eqn:gradient_y}
\end{equation} 
in which $\frac{\partial \mathcal{L}}{\partial \bm{\mathrm{Y}}} \in \mathbb{R}^{ (B_{c}/r + B_{p}) \times C\cdot r \times D}$. $F_{S,g}^{-1}$ and $F_{A,g}$ stands for the backward operation of $F_{S}^{-1}$ and $F_{A}$, respectively. The backward operation of $F_{S,g}^{-1}$ is the reverse-direction-reshape of a given gradient: $F_{S,g}^{-1}(\frac{\partial \mathcal{L}}{\partial \bm{\mathrm{Y}}_{B_c}}; r) = F_{S}(\frac{\partial \mathcal{L}}{\partial \bm{\mathrm{Y}}_{B_c}}; r):\mathbb{R}^{B_c \times C \times D}\rightarrow\mathbb{R}^{B{_c}/r \times C\cdot r \times D}$.
In the case of $F_{A}$, the backward operation becomes $F_{A,g}(\frac{\partial \mathcal{L}}{\partial \bm{\mathrm{Y}}_{B_c}};r) = ((\frac{\partial \mathcal{L}}{\partial \bm{\mathrm{Y}}_{B_p}}){_1}/r;\cdots;(\frac{\partial \mathcal{L}}{\partial \bm{\mathrm{Y}}_{B_p}}){_r}/r):\mathbb{R}^{B_p \times C \times D}\rightarrow\mathbb{R}^{B_p \times C\cdot r \times D }$, where distributes and expands the input gradient divided by $r$, for $r$ times.
As a result, a gradient of $\beta$ and $\gamma$ becomes:
\begin{equation}
\vspace{-.1in}
\frac{\partial \mathcal{L}}{\partial \gamma} = F_{P, g}\bigg(\sum^{(B{_c}/r + B_p), D}_{b,d=1} \frac{\partial \mathcal{L}}{\partial \bm{\mathrm{Y}}_{b,d}} \odot \tilde{\bm{{\mathrm{H}}}}_{b,d};r\bigg) \label{eqn:tbbn_gradient_gamma}\ \ \
\end{equation} 
\begin{equation}
\frac{\partial \mathcal{L}}{\partial \beta} = F_{P, g}\bigg(\sum^{(B{_c}/r + B_p), D}_{b,d=1} \frac{\partial \mathcal{L}}{\partial \bm{\mathrm{Y}}_{b,d}};r\bigg), \label{eqn:tbbn_gradient_beta}
\end{equation}
in which $F_{P, g}$ denotes the backward operation of $F_{P}$, which is the sum of gradients over $r$: $F_{P, g}(\bm g ;r)=\sum_{i=1}^{r} ((\bm g_i); \cdots; (\bm g_r)):\mathbb{R}^{C\cdot r} \rightarrow \mathbb{R}^{C}$, and
$\frac{\partial \mathcal{L}}{\partial \bm{\mathrm{Y}}_{b,d}}  \in\mathbb{R}^{C}$ and $\tilde{\bm{{\mathrm{H}}}}_{b,d} \in\mathbb{R}^{C}$. 
Note that the core differences with the gradients of the original BN shown in (\ref{eqn:gradient_gamma_beta}): First, the summed up gradients are configured to be task-balanced. Second, in the case of the gradient of $\gamma$, the task-balanced batch $\tilde{\bm{{\mathrm{H}}}}$, normalized by the task-balanced mean and variance, is used to get the gradient.  
As a result, $\gamma$ and $\beta$ can be trained in a more task-balanced way, so that they can do their part of BN at test time where the entire task's test data is given jointly.

\csm{Note that both the pseudo code and implementation details of TBBN are proposed in supplementary materials.}

\section{Experimental Results}
\vspace{-.05in}
\noindent\textbf{Datasets and Evaluation.} \ \
We mainly evaluate our method on the offline CIL settings leveraging CIFAR-100~\cite{(cifar)krizhevsky2009learning} and ImageNet-100~\cite{(imagenet)deng2009imagenet} datasets: for each dataset, we divide 100 classes into 10 disjoint sets, and consider each set to constitute a single task (\textit{i.e.} $T = 10$ and $m = 10$).
Following previous work~\cite{(cil_survey)masana2020class, (cl_survey)delange2021continual, (cpr)cha2020cpr}, we evaluate all normalization methods based on four different metrics, such as {final accuracy} ($A_f$), {average accuracy} ($A_a$), {forgetting measure}($F$), and {learning accuracy}($A_l$).
The detailed explanation for the metrics can be found in the supplementary materials.


\noindent\textbf{Baselines.} \ \
For backbone architectures, we follow the offline CIL setup proposed in \cite{(cil_survey)masana2020class} using ResNet-32~\cite{(resnet)he2016deep} for CIFAR-100~\cite{(cifar)krizhevsky2009learning} and ResNet-18 for ImageNet-100~\cite{(imagenet)deng2009imagenet}.
In addition, we use exemplar memory with size $|\mathcal{M}| = 2000$ and incrementally update $\mathcal{M}$ via a class-balanced random sampling strategy after each task.
To test our method under various offline CIL algorithms, we consider EEIL~\cite{(eeil)castro2018end} and LUCIR~\cite{(lucir)hou2019learning}, for which implementations are available in the benchmark environment provided by \cite{(cil_survey)masana2020class}.
We also run SS-IL~\cite{(ss-il)ahn2021ss}
, PODNet~\cite{(podnet)douillard2020podnet} and AFC~\cite{(AFC)kang2022class}, which are considered state-of-the-art in offline CIL, using their official code.
Finetuning (FT) stands for a naive baseline which only utilizes the exemplar memory without using any explicit methods.
For comparison across different normalization approaches, we simply replace all BN layers in the backbone model with one of the following: Instance Normalization (IN)~\cite{(IN)ulyanov2016instance}, Group Normalization (GN)~\cite{(GroupBN)wu2018group}, Switch Normalization (SN)~\cite{(SN)luo2019switchable}, Batch Renormalization (BRN)~\cite{(BRN)ioffe2017batch}, Continual Normalization (CN)~\cite{(CN)pham2022continual}, and our proposed TBBN. We use official implementations for each baseline method published by the respective authors. Note that CN and GN require a hyperparameter $G$ that sets the group size during feature-level normalization. For our experiments, we use $G = 16$ when training ResNet-32 on CIFAR-100 and $G = 32$ when training ResNet-18 on ImageNet-100.
Further details on experimental settings are in the supplementary material.

\subsection{\csm{Quantitative Results}}
\vspace{-.05in}
\noindent\textbf{Effect of normalization layers in offline CIL.}  \ \
Table~\ref{table:norm_layers} shows experimental results for FT on 10-task offline CIL setups on CIFAR-100 and ImageNet-100. 
First, we find that normalization layers previously developed for single task supervised learning (e.g., IN~\cite{(IN)ulyanov2016instance}, GN~\cite{(GroupBN)wu2018group}, SN~\cite{(SN)luo2019switchable} and BRN~\cite{(BRN)ioffe2017batch}) are not effective in offline CIL scenarios compared to BN.
Both IN and GN especially show notable performance degradation with respect to all four metrics.
Furthermore, CN~\cite{(CN)pham2022continual} does not lead to significant improvement, which is in contrast to the performance gain previously demonstrated in online CIL by the original paper. This shows that the benefit of adding the channel-level GN to BN is obscured when na\"ive BN is exposed to imbalanced mini-batches for multiple epochs.
On the other hand, we confirm that simply replacing BN with TBBN increases both $A_f$ and $A_a$ in both datasets without any additional hyperparameters. We find that the performance gain mainly comes from the increased stability as well as plasticity as we see improvements in both $F$ and $A_l$.
TBBN especially shows significantly less forgetting, supporting our insight that preventing sample statistics from becoming biased to the current task via task-balanced normalization effectively retains information from past tasks leading to a better accuracy.


\begin{table}[h]
\caption{Experimental results for FT with varying normalization layers on CIFAR-100 and ImageNet-100. Bold indicates the best performance in each metric.}
\label{table:norm_layers}
\centering
\smallskip\noindent
\vspace{-.2in}
\resizebox{\linewidth}{!}{
    \begin{tabular}{c|cccc|cccc}
    \toprule
    \multirow{2}{*}{Method} & \multicolumn{4}{c|}{CIFAR-100 w/ ResNet-32} & \multicolumn{4}{c}{ImageNet-100 w/ ResNet-18}\\
     & $A_f(\uparrow)$ & $A_a(\uparrow)$ & $F(\downarrow)$ & $A_l(\uparrow)$ & $A_f(\uparrow)$ & $A_a(\uparrow)$ & $F(\downarrow)$ & $A_l(\uparrow)$\\
    \midrule
    BN & 35.41 & 53.88 & 43.48 & 78.79 & 39.40 & 59.60 & 48.02 & 87.42 \\ 
    IN & 31.72 & 46.84 & 47.72 & 79.44 & 33.45 & 53.59 & 50.69 & 84.66 \\ 
    GN & 31.26 & 44.53 & 44.01 & 75.27 & 28.83 & 47.79 & 49.79 & 79.19 \\
    SN & 36.29 & 53.64 & 42.91 & 79.20 & 39.45 & 59.55 & 48.04 & 87.79 \\ 
    BRN & 36.08 & 52.58 & 44.34 & 80.42 & 37.49 & 57.77 & 48.13 & 86.57 \\
    \midrule
    CN & 35.06 & 54.43 & 43.82 & 80.64 & 41.96 & 60.02 & 45.32 & 87.28 \\ 
    CN$^{*}$ & 36.05 & 54.18 & 44.81 & \textbf{80.85} & 40.46 & 59.03  & 46.74 & 87.20 \\
    \midrule
    TBBN & \textbf{38.46} & \textbf{56.17} & \textbf{41.90} & 80.36 & \textbf{43.20} & \textbf{61.69} & \textbf{43.62} & \textbf{87.82} \\ 
    \bottomrule
    \end{tabular}
}
\end{table}
\vspace{-.1in}

\noindent\textbf{Applying TBBN to other CIL baselines.} \ \
We also confirm that the advantage of TBBN is extended to other CIL methods.
We consider the same 10-task offline CIL setup from CIFAR-100 and ImageNet-100, and run a total of six different CIL methods while simply replacing the BN layers in the model with either CN or TBBN.

Table \ref{table:cil_baselines} shows the corresponding results.
First, the most recently proposed CIL methods, SSIL and AFC, achieve the best performance in terms of final accuracy $A_f$. Especially in ImageNet-100, they surpass all other baselines in both $A_f$ and $A_a$, as previously reported in the original papers. 
Second, replacing BN with CN does not show consistent improvement across datasets and CIL methods.
For instance, while EEIL+CN produces a considerable performance gain in ImageNet-100, CN fails to outperform BN when combined with more recent CIL algorithms such as PODNet and AFC.
On the other hand, our TBBN consistently improves performance in $A_f$ and $A_a$ except when applied on PODNet in CIFAR-100. Following the observation in the previous experiment, we again confirm that the performance gain of TBBN is mostly due to reduction in $F$ (\textit{i.e.} enhanced stability) while maintaining or increasing $A_l$ (\textit{i.e.} enhanced plasticity). In all, applying TBBN to the SOTA baselines achieves new SOTA performance in both datasets. We plotted a graph of these experiments in the supplementary materials, and we could confirm that TBBN enhances the average accuracy at every step of training compared to BN, while CN shows performance loss for some PODNet and AFC.

\begin{table}[h]
\caption{Experimental results for various representative offline CIL methods. Bold indicates the best performance in each metric.}
\vspace{-.1in}
\centering
\smallskip\noindent
\resizebox{\linewidth}{!}{
\begin{tabular}{cc|cccc|cccc}
\toprule
\multicolumn{2}{c|}{\multirow{2}{*}{Method}} & \multicolumn{4}{c|}{CIFAR-100 w/ ResNet-32} & \multicolumn{4}{c}{ImageNet-100 w/ ResNet-18}\\
& & $A_f(\uparrow)$ & $A_a(\uparrow)$ & $F(\downarrow)$ & $A_l(\uparrow)$ & $A_f(\uparrow)$ & $A_a(\uparrow)$ & $F(\downarrow)$ & $A_l(\uparrow)$\\
\midrule
\multirow{3}{*}{FT} & +BN & 35.41 & 53.88 & 43.38 & 78.79 & 39.40 & 59.60 & 48.02 & 87.42 \\
& +CN & 35.06 & 54.43 & 43.82 &\textbf{80.64} & 41.96 & 60.02 & 45.32 & 87.28 \\ 
& +TBBN & \textbf{38.46} & \textbf{56.17} & \textbf{41.90} & 80.36 & \textbf{43.20} & \textbf{61.69} & \textbf{43.62} & \textbf{87.82} \\
\midrule
\multirow{3}{*}{EEIL} & +BN & 39.82 & 55.25 & 39.40 & 79.22 & 40.06 & 61.15 & 47.78 & \textbf{87.84} \\
& +CN & 39.98 & 55.09 & 39.31 & 79.29 & 42.48 & 61.43 & 44.44 & 86.92 \\ 
& +TBBN & \textbf{41.93} & \textbf{57.53} & \textbf{37.80} & \textbf{79.93} & \textbf{45.18} & \textbf{63.48} & \textbf{42.66} & \textbf{87.84} \\ 
\midrule
\multirow{3}{*}{LUCIR} & +BN & 38.06 & 54.20 & 32.35 & 70.41 & 42.26 & 63.82 & 41.68 &\textbf{83.94} \\ 
& +CN & 38.07 & 55.60 & 33.78 & \textbf{71.85} & 40.44 & 61.44 & 42.04 & 83.48 \\ 
& +TBBN & \textbf{41.45} & \textbf{56.13} & \textbf{29.23} & 70.68 & \textbf{43.72} &\textbf{64.36} & \textbf{40.18}& 83.90 \\
\midrule
\multirow{3}{*}{PODNet} & +BN & \textbf{38.10} & 52.95 & 14.70 & \textbf{52.58} &  49.05 & 65.41 & 22.40 & \textbf{69.99}\\ 
& +CN & 34.80 & 50.26 & 15.69 & 50.52 & 46.20   & 62.91 & 23.66 & 68.50\\ 
& +TBBN & 37.90 &\textbf{52.98} & \textbf{13.90} & 51.78 & \textbf{49.30}   & \textbf{65.70} & \textbf{21.85} & 69.76\\ 
\midrule
\multirow{3}{*}{SSIL} & +BN & 41.34 & 53.00 & 15.64 & 56.02 & 49.56 & 65.79 & 21.20 & 69.94 \\ 
& +CN & 40.74 & 52.38 & \textbf{14.60} & 54.44 & 50.58 & 64.81 &\textbf{ 18.56} & 65.04 \\ 
& +TBBN &\textbf{43.80} & \textbf{54.28} & 15.12 & \textbf{59.37} & \textbf{51.30} & \textbf{66.51} & 19.58 & \textbf{70.64} \\
\midrule
\multirow{3}{*}{AFC} & +BN & 39.90 & 53.93 & 33.17 & 73.10 & 52.50 & 67.53 & 19.70 & 72.22 \\ 
& +CN & 37.50 & 51.16 & 33.40 & 70.94 & 48.00 & 65.21 & 20.10 & 70.68 \\ 
& +TBBN & \textbf{41.30} & \textbf{57.31} & \textbf{32.89} & \textbf{73.57}& \textbf{54.00} &\textbf{ 68.68} & \textbf{19.00} & \textbf{73.22}\\ 
\bottomrule
\end{tabular}
}
\vspace{-.15in}
\label{table:cil_baselines}
\end{table}

\noindent\textbf{Applying TBBN to other architectures.}
To check the applicability of TBBN to other architectures, we selected four CNN-based architectures (\textit{e.g.}, ResNet-34~\cite{(resnet)he2016deep}, ShuffleNet-V2~\cite{(shufflenet)zhang2018shufflenet}, MobileNet-V2~\cite{(mobilenetv2)sandler2018mobilenetv2}, and MnasNet (x0.5)~\cite{(mnasnet)tan2019mnasnet}) which contain the BN layer as a default, and conducted experiments by replacing BN with TBBN in the 10 tasks scenario using the ImageNet-100 dataset.
Table \ref{table:architecture} demonstrates that our TBBN can be successfully applied to various types of architecture, increasing both A$_f$ and A$_a$ concurrently.

\begin{table}[h]
\caption{Experimental results with varying backbone architectures on ImageNet-100. Bold indicates the best performance.}
\vspace{-.2in}
\centering
\smallskip\noindent
\resizebox{\linewidth}{!}{
\begin{tabular}{cc|cc|cc|cc|cc}
\toprule
\multicolumn{2}{c|}{\multirow{2}{*}{Method}} & \multicolumn{2}{c|}{ResNet-34} & \multicolumn{2}{c|}{ShuffleNet-V2} & \multicolumn{2}{c|}{MobileNet-v2} & \multicolumn{2}{c}{MnasNet (x0.5)}\\
& & $A_f(\uparrow)$ & $A_a(\uparrow)$ & $A_f(\uparrow)$ & $A_a(\uparrow)$ & $A_f(\uparrow)$ & $A_a(\uparrow)$ & $A_f(\uparrow)$ & $A_a(\uparrow)$\\
\midrule 
\multirow{2}{*}{FT} & +BN & 41.38 & 62.00 & 34.74 & 55.37 & 38.86 & 57.43 & 36.24 & 54.96\\
& +TBBN & \textbf{46.72} & \textbf{64.76} & \textbf{35.82} & \textbf{58.17} & \textbf{42.66} & \textbf{60.41} & \textbf{40.94} & \textbf{57.09}\\
\midrule
\multirow{2}{*}{SSIL} & +BN & 51.18 & 65.27 & 43.86 & 59.69 & 47.72 & 62.25 & 45.10 & 59.81\\
& +TBBN & \textbf{52.30} & \textbf{66.36} & \textbf{44.72} & \textbf{60.07} & \textbf{48.62} & \textbf{63.37} & \textbf{47.06} & \textbf{60.24}\\
\bottomrule
\end{tabular}
}
\vspace{-.1in}
\label{table:architecture}
\end{table}
\noindent\textbf{Results on dissimilar tasks.} \ \
Several benchmark datasets (\textit{e.g.}, CIFAR-10/-100 and ImageNet dataset) have been used to evaluate a CIL method. The CIL scenario made from random class orderings of these datasets has \textit{small domain shifts} due to high similarity between tasks~\cite{(cil_survey)masana2020class}.
In this regard, we believe that the CIL scenario using CIFAR-100 and ImageNet-100 datasets is not the best setting to check whether the proposed normalization layer is suitable for the CIL.
Therefore, we design the CIL scenario, which has \textit{large domain shifts}, with five distinct datasets such as CIFAR-10~\cite{(cifar)krizhevsky2009learning}, SVHN~\cite{(svhn)netzer2011reading}, STL-10~\cite{(stl10)coates2011analysis}, MNIST~\cite{(mnist)deng2012mnist} and FashionMNIST~\cite{(fmnist)xiao2017fashion}.
We divide 10 classes of each dataset into 2 tasks and construct the CIL scenario consisting of 10 tasks ($2$ tasks $\times$ 5 datasets). Note that task ordering is randomly shuffled by random seed and more details for experimental settings are proposed in the supplementary materials.

In Table \ref{table:dissimilar}, we observe that applying CN is sometimes harmful and only achieves minor improvement for FT and LUCIR.
On the other hand, we again confirm that TBBN constantly makes enhancement of $A_{f}$ and $A_{a}$ by diminishing $F$ but well maintaining $A_l$, for all methods. This result underscores the robustness of our TBBN even for the difficult CIL scenario consisting of dissimilar tasks.

\begin{table}[h]
\vspace{-.1in}
\caption{Experimental results with dissimiar tasks using ResNet-18. Bold indicates the best performance in each metric.}
\vspace{-.15in}
\centering
\smallskip\noindent
\resizebox{.7\linewidth}{!}{
\begin{tabular}{cc|cccc}
\toprule
\multicolumn{2}{c|}{Method} & $A_f(\uparrow)$ & $A_a(\uparrow)$ & $F(\downarrow)$ & $A_l(\uparrow)$ \\
\midrule
\multirow{3}{*}{FT} & +BN & 59.60 & 65.85 & 31.48 & 91.08 \\ 
& +CN & 61.41 & 71.00 & 29.90 & 91.40 \\ 
& +TBBN & \textbf{63.42} & \textbf{71.32} & \textbf{28.05} & \textbf{91.47} \\ 
\midrule
\multirow{3}{*}{EEIL} & +BN & 62.04 & 72.28 & 29.46 & 91.50 \\ 
& +CN & 61.73 & 72.40 & 29.27  & 91.00 \\ 
& +TBBN & \textbf{64.88} & \textbf{72.80} & \textbf{26.67} & \textbf{91.55} \\ 
\midrule
\multirow{3}{*}{LUCIR} & +BN & 62.42 & 72.91 & 28.09 & \textbf{90.35} \\
& +CN & 62.99 & 72.81 & \textbf{26.70} & 89.49 \\
& +TBBN & \textbf{63.64} & \textbf{73.72} & 26.73 & 90.23 \\ 
\midrule
\multirow{3}{*}{SSIL} & +BN & 66.15 & 70.62 & 10.13 & 72.59 \\
& +CN & 65.64 & 70.56 & \textbf{9.16} & 69.66 \\
& +TBBN & \textbf{67.00} & \textbf{71.77} & 9.74 & \textbf{72.86} \\ 
\bottomrule
\end{tabular}
}
\vspace{-.1in}
\label{table:dissimilar}
\end{table}


\subsection{Qualitative Analysis}
\begin{figure}[h]
\vspace{-.05in}
    \centering
{\includegraphics[width=0.72\linewidth]{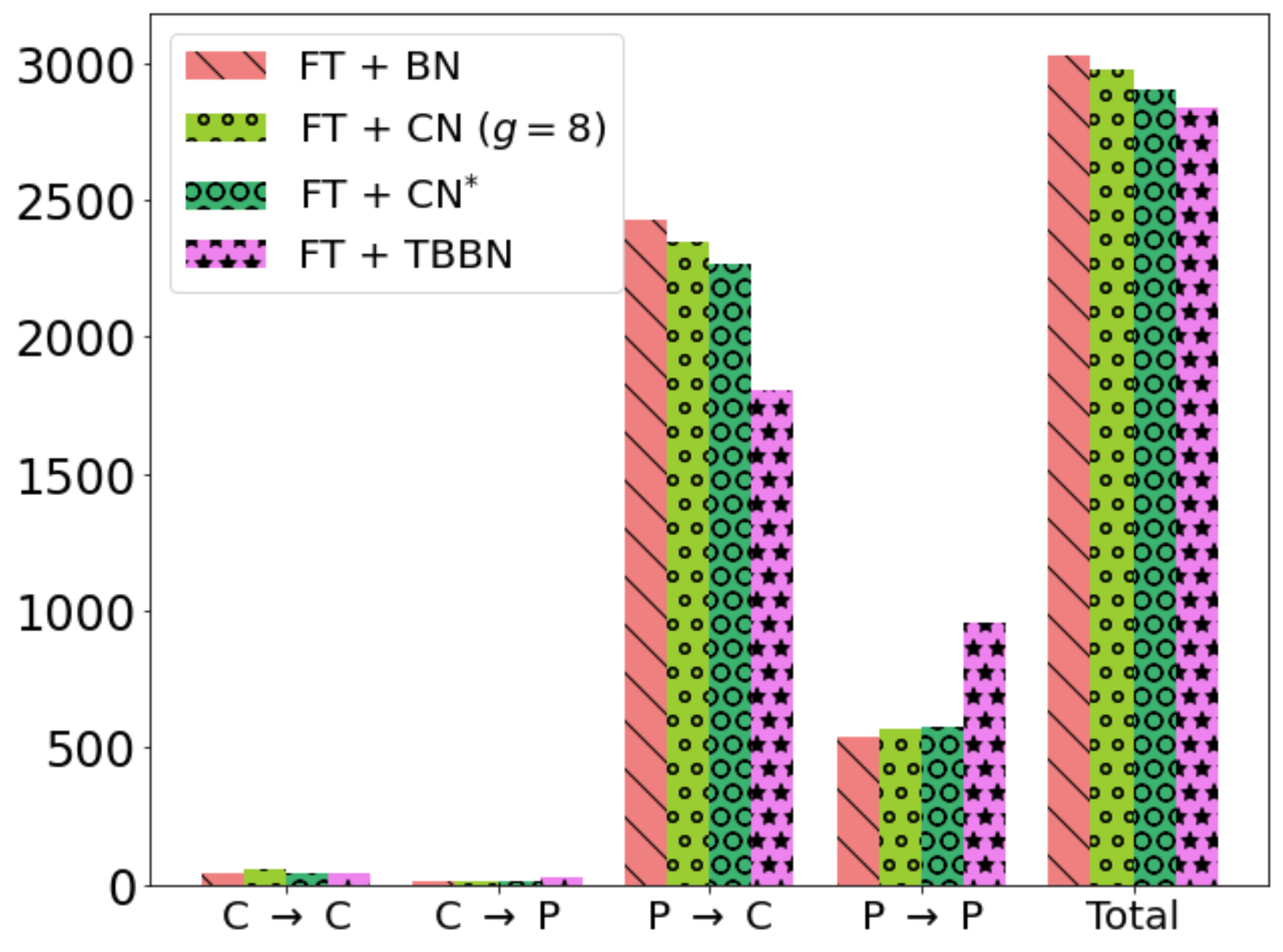}}
\vspace{-.15in}
\caption{\csm{Analysis for four types of misclassification errors ($y$-axis) after training the model until the final task ($t=10$). }}
\vspace{-.1in}
\label{figure:forgetting}
 \end{figure}

\vspace{-.05in}
\noindent\textbf{Reduction in biased predictions.} \ \
For a more detailed analysis for the effect of TBBN, we investigated a type of misclassification, by following the experiment proposed in \cite{(il2m)belouadah2019il2m}, and the analysis results for finetuning with the ImageNet-100 dataset (10-tasks) are reported in Figure \ref{figure:forgetting}.
This figure presents the number of four types of misclassification.
For example, C $\rightarrow$ P stands for the number of misclassification current task's data to another class in previous tasks ($t = 1, \dots, 9)$.
Among four types, P $\rightarrow$ C, which is the number of misclassification of the previous task's data toward the current task's class, is called as \textit{biased prediction}, knowing as the major cause of degrading performance in CIL~\cite{(bic)wu2019large,(ss-il)ahn2021ss}.
As already reported in previous works~\cite{(il2m)belouadah2019il2m, (bic)wu2019large}, FT + BN seriously suffers from the biased prediction, resulting in the largest number of total misclassification.
In the case of both CN variants and TBBN, we confirm that the gain of reducing the total number of misclassifications comes from diminishing the number of biased predictions (P $\rightarrow$ C),
Especially, our TBBN significantly alleviates the biased prediction than CN variants. As a result, FT + TBBN achieves the lowest number of total misclassifications despite the increase of P $\rightarrow$ P caused by the reduced biased prediction.

\noindent\textbf{Ablation studies.} \ \
Table \ref{table:ablation_study} presents ablation study for TBBN using the CIFAR-100 dataset (10 tasks).
First, both Case 1 and Case 2 represent the ablation study for task-balanced mean and variance. We confirm that only calculating task-balanced running mean $\mu$ and variance $\sigma^2$ in (\ref{eqn:moving_average}) for test phase (Case 1) or using a task-balanced running mean $\tilde{\sigma}$ and variance $\tilde{\sigma}^2$ for training phase (Case 2) affect final performance negatively, due to mismatch of values ($\mu$, $\sigma^2$) for normalization in training and test phases.
Case 3 shows the experimental result where calculating task-balanced mean and variance is only applied. However, as we already confirmed in Figure \ref{figure:motivation_experiments} of the Section 4.1, only considering task-balanced mean and variance cannot make a difference in performance with the original BN. The opposite case (Case 4) where $\gamma$ and $\beta$ are trained in a task-balanced manner also shows a similar tendency, demonstrating the performance enhancement by TBBN can be obtained only when the proposed components are used together.

Additionally, we compare GPU memory cost, training time and inference time of each case. Because of both reshape and repeat operations consisting of TBBN, TBBN requires maximum $\times1.6$ of GPU memory and takes more $\times1.3$ of training time than the original BN. However, note that there is no additional cost at test time because TBBN works to be identical to the original BN.

\begin{table}[h]
\vspace{-.1in}
\caption{Ablation study using CIFAR-100 with ResNet-32.}
\vspace{-.1in}
\centering
\smallskip\noindent
\resizebox{\linewidth}{!}{
\begin{tabular}{c|c|c|c|cc|c|c|c}
\toprule
\multirow{3}{*}{Method} 
& \multicolumn{2}{c|}{Task-balanced}
& \multirow{3}{*}{\begin{tabular}[c]{@{}c@{}}Task-balanced\\ $\gamma$ and $\beta$?\end{tabular}}
& 
& 
& \multirow{3}{*}{\begin{tabular}[c]{@{}c@{}}Peak\\ Memory\end{tabular}}
& \multirow{3}{*}{\begin{tabular}[c]{@{}c@{}}Train Time\\ (total)\end{tabular}}
& \multirow{3}{*}{\begin{tabular}[c]{@{}c@{}}Test Time\\ (per batch)\end{tabular}}\\
& \multicolumn{2}{c|}{$\mu$ and $\sigma^2$?} & & $A_f(\uparrow)$ & $A_a(\uparrow)$ & & & \\
& Train & Test &  & & & & & \\
\midrule
TBBN & \cmark & \cmark & \cmark & \textbf{38.46} & \textbf{56.17} & \multirow{5}{*}{\begin{tabular}[c]{@{}c@{}}2525MiB \\ $(\times 1.6)$\end{tabular}} & \multirow{5}{*}{\begin{tabular}[c]{@{}c@{}}6.0h \\ $(\times 1.3)$\end{tabular}} & \multirow{5}{*}{\begin{tabular}[c]{@{}c@{}}0.6s \\ $(\times 1.0)$\end{tabular}}\\
Case 1 & \xmark & \cmark & \cmark & 29.78 & 46.99 & & &\\
Case 2 & \cmark & \xmark & \cmark & 0.93 & 9.69 & & &\\
Case 3 & \cmark & \cmark & \xmark & 35.45 & 53.15 & & &\\
Case 4 & \xmark & \xmark & \cmark & 36.21 & 53.25 & & &\\
\midrule
BN & \xmark & \xmark & \xmark & 35.71 & 53.88 & 1545MiB & 4.6h & 0.6s\\
\bottomrule
\end{tabular}
}
\vspace{-.15in}
\label{table:ablation_study}
\end{table}

\section{Concluding Remarks}

We propose a simple but effective method, called Task-Balance Batch Normalization, for exemplar-based CIL. Starting from \csm{an analysis of} the problem of the original BN, \csm{namely}, the biased mean and variance calculation toward the current task, we devise \csm{a} novel method to calculate the task-balanced mean and variance for normalization. \csm{Furthermore}, we propose a method for the task-balanced training of parameters for affine transformation. From extensive experiments with CIFAR-100,  ImageNet-100 and dissimilar tasks, we demonstrate that our TBBN can be easily applied to various existing CIL methods, \csm{further} improving their performance. 
\section{Acknowledgement}
This work was partly done while Sungmin Cha did a research internship at Advanced ML Lab, LG AI Research. The work was also supported in part by NRF grant [NRF-2021R1A2C2007884], IITP grant [No.2021-0-01343, No.2021-0-02068, No.2022-0-00113, No.2022-0-00959] funded by the Korean government (MSIT), and SNU-LG AI Research Center.


{\small
\bibliographystyle{ieee_fullname}
\bibliography{example_paper}
}

\end{document}


\title{Supplementary Materials for \\ Rebalancing Batch Normalization for \\ Exemplar-based Class-Incremental Learning}

\author{
Sungmin Cha\textsuperscript{\rm 1},\  Sungjun Cho\textsuperscript{\rm 2},\  Dasol Hwang\textsuperscript{\rm 2},\  Sunwon Hong\textsuperscript{\rm 1}, \  Moontae Lee\textsuperscript{\rm 2,3},\  and Taesup Moon\textsuperscript{\rm 1,4,5}\thanks{Corresponding author (E-mail: \texttt{tsmoon@snu.ac.kr})}\vspace{.05in} \\
  \textsuperscript{\rm 1}Department of ECE, Seoul National University\ \
  \textsuperscript{\rm 2}LG AI Research \ \
  \textsuperscript{\rm 3}University of Illinois Chicago \\
  \textsuperscript{\rm 4}ASRI / INMC / IPAI / AIIS, Seoul National University\ \ 
  \textsuperscript{\rm 5}SNU-LG AI Research Center \\
  \texttt{sungmin.cha@snu.ac.kr},\ \ \texttt{\{sungjun.cho, dasol.hwang\}@lgresearch.ai}, \\ \texttt{zghdtnsz96@snu.ac.kr},\ \ \texttt{moontae.lee@lgresearch.ai}, \texttt{tsmoon@snu.ac.kr}
}

\maketitle

\newpage

\begin{figure*}[t]
\centering 
\subfigure[CIFAR-100 (10 Tasks)]
{\includegraphics[width=0.48\linewidth]{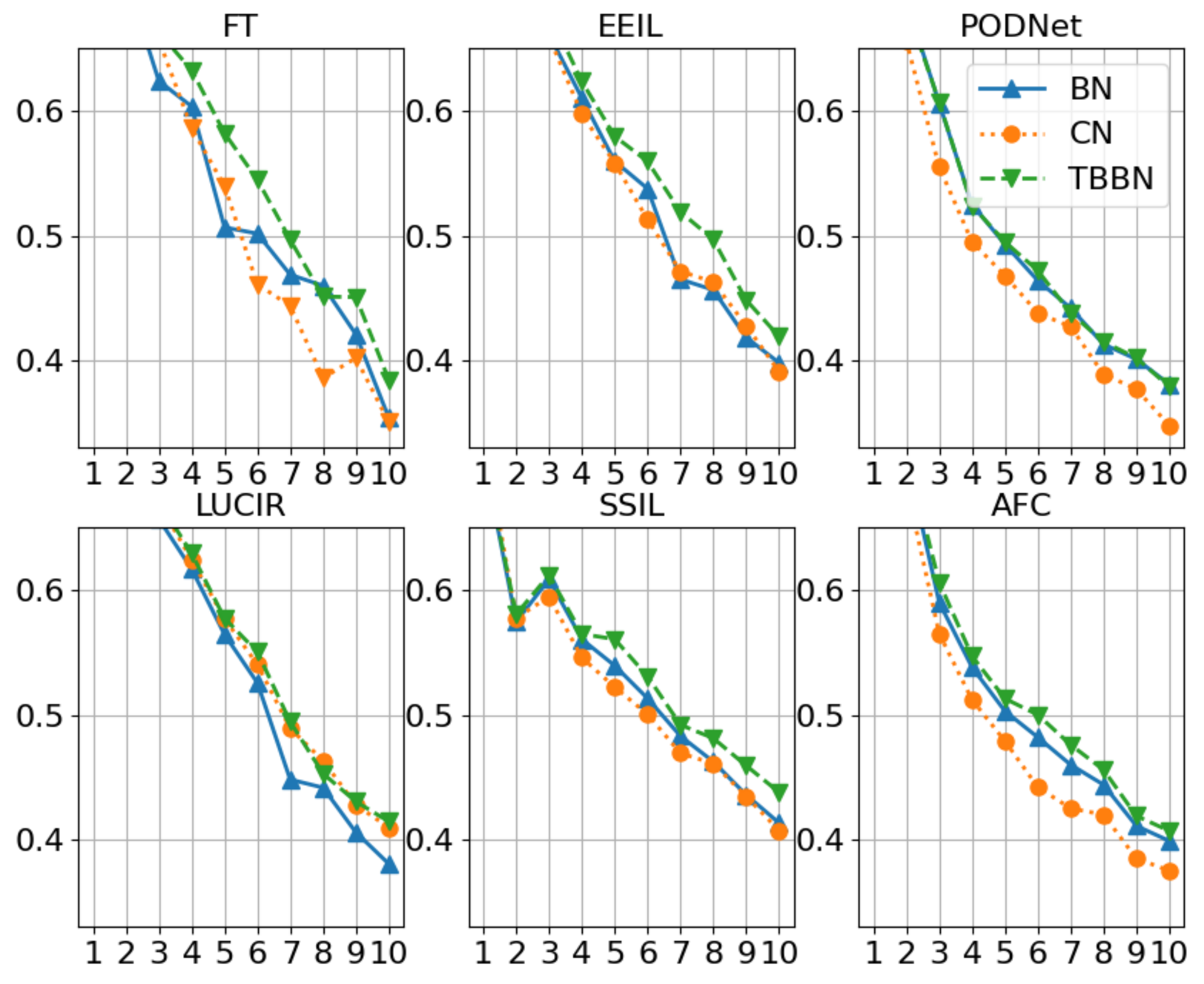}
\label{figure:whitebox_untargeted_eps}}
\subfigure[ImageNet-100 (10 Tasks)]
{\includegraphics[width=0.48\linewidth]{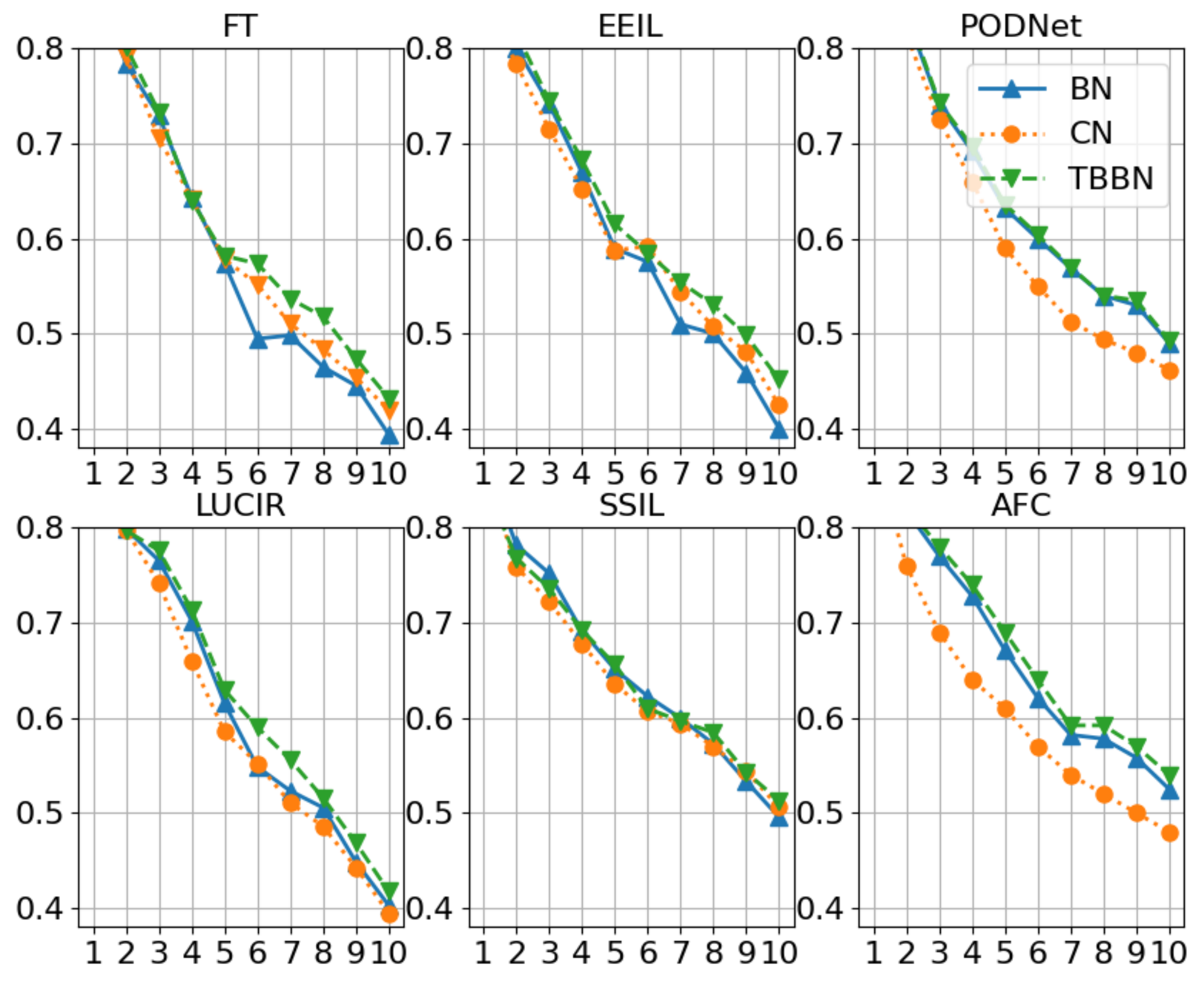}
\label{figure:whitebox_untargeted_iter}}
\vspace{-.15in}
\caption{Accuracy curves during training with various CIL algorithms. X-axes show the number of tasks $t$, and the Y-axes show accuracies averaged across already seen $t$ number of tasks. Replacing the BN layers (solid lines) with TBBN (dotted lines) leads to consistently better accuracies throughout training.}
\label{figure:baselines_figure}
\vspace{-.2in}
 \end{figure*}


\section{Implementation details of TBBN}\label{section:sec1.2}
There are some considerations to implementing our TBBN in exemplar-based CIL.
Firstly, in order to use TBBN, the values for $B_c$, $B_p$, and information about task changes are required, and the ratio $\frac{B_{c}}{B_{p}}$ must be an integer. However, we believe that this information is readily available and adjustable in a general offline CIL scenario, as already shown in \cite{(bic)wu2019large,(ss-il)ahn2021ss}.
Secondly, it should be noted that not all adaptively determined values for $r$ can reshape a given feature map. For instance, when $\frac{B_{c}}{r}$ is not an integer, the tensor reshape operation $F_{RS}$ cannot be applied. We overcome this limitation by using a simple rule for determining $r$. After calculating $r$ using Equation (6) (as presented in the manuscript) at the beginning of each task training, we set a feasible $r^{*}$ using the following rule:
\begin{equation}
  r^{*}=\left\{
  \begin{array}{@{}ll@{}}
    r, & \text{if}\ r \in CD(B_c, B_p) \\
    M(B_c, B_p, r), & \text{otherwise}
  \end{array}\right.
\end{equation} 
where $M(B_c, B_p, r) = \max \{ \hat{r} : \hat{r} \in CD(B_c, B_p) \wedge \hat{r} < r\}$ and $CD(\cdot, \cdot)$ denotes a set of common divisors between two values.
Although $r^{*}$ is not the exact optimal value for our TBBN, we already experimentally confirmed that using $r^{*}$ is also effective for most CIL experiments in the manuscript.
Finally, it should be noted that there is no difference between the original BN and our TBBN in the test phase because TBBN also maintains ${\mu}, {\sigma}^2, \gamma, \beta \in \mathbb{R}^C$ during the training phase.

\section{Evaluation Metrics}\label{section:sec2}

Let $a_{t,i} \in [0,1]$ denote the accuracy on the test set of task $i$ after training on the first $t$ tasks.
The \textit{final accuracy} $A_f = \frac{1}{T}\sum_{i=1}^{T} a_{T,i}$ measures the classification accuracy of the model at the end of training averaged across all tasks, and the \textit{average accuracy} $A_a = \frac{1}{T}\sum_{t=1}^{T} \left(\frac{1}{t}\sum_{i=1}^{t} a_{t,i}\right)$ measures the average accuracy until task $T$. Note that while these two metrics gauge the discriminative performance of the CIL pipeline, they do not reflect the stability-plasticity aspect, for which the following two metrics have been designed.
The \textit{forgetting measure} $F = \frac{1}{T} \sum_{i=1}^{T} \max_{t\in[i+1,T]}(a_{i,i} - a_{t,i})$ proposed by \cite{(rwalk)chaudhry2018riemannian} measures the degree of forgetting by averaging the maximum decrease in accuracy of all tasks throughout the course of training. 
Lastly, the \textit{learning accuracy} $A_l = \frac{1}{T}\sum_{i=1}^{T} a_{i,i}$ proposed by \cite{(la)riemer2018learning} measures the plasticity of the model by averaging the accuracy of each task immediately after training on that task. We report all measurements averaged across three runs with different seeds.

\section{Additional Experimental Results}\label{section:sec3}

\subsection{Accuracy curves}\label{section:sec3.1} \ \ 
To visualize the task accuracies during training, Figure~\ref{figure:baselines_figure} displays the average classification accuracy across all previously seen tasks throughout the training process ($A_f$ after each task). Our observations show that TBBN improves the average accuracy at every step of training compared to BN, whereas CN exhibits fluctuations that result in performance degradation when applied to AFC and PODNet.

\subsection{Experimental results for making a balanced batch with data augmentation}\label{section:sec3.2}

To confirm the novelty of TBBN, we conducted an experiment for making a balanced batch with data augmentation for sampled data in the exemplar memory.
We selected powerful augmentation methods which are widely used for self-supervised contrastive learning, consisting of \textit{RandomResizedCrop}, \textit{RandomHorizontalFlip}, \textit{ColorJitter} and \textit{RandomGrayscale}.
To make the balanced batch at each $t(>1)$-th task's training time, we augmented each data sample in the exemplar memory for $(t-1) \times 3$ times. 
This is because we set the ratio between the data points from the current task and those from the exemplar memory to $3 : 1$, and the sampled batch from the exemplar memory always contains data from $t-1$ numbers of the previous task.
Table \ref{table:ft_balaug} shows the average result on CIFAR-100 for FT with balanced augmentation (FT + BalAug) with \textit{ordinary} BN for 3 seeds.
We observe that this baseline does not bring a positive performance gain, compared to the FT+BN in (Table 1, manuscript).
We believe that the FT+BalAug has two limitations: 1) Despite the augmentation, the model ends up over-fitting to the samples in the exemplar memory due to the scarcity of data, and 2) $t \times 3$ times of augmentation for previous task's data enlarge the size of mini-batch for each task, causing the computation and memory cost increase during training. 
We believe this result further demonstrates the effectiveness of our TBBN for the exemplar-based CIL. 

\begin{table}[h]
\caption{Experimental results of FT + BalAug (with BN).}
\centering
\smallskip\noindent
\resizebox{.8\linewidth}{!}{
\begin{tabular}{|c||c|}
\hline
Acc / FM / LA & CIFAR-100 ($T=10$)                          \\ \hline \hline
FT + BalAug & 25.54 / 42.76 / 68.27 \\ \hline
\end{tabular}
}
\vspace{-.1in}
\label{table:ft_balaug}
\end{table}

\subsection{Experimental results for other CIL protocol (using base task)}\label{section:sec3.3}

\begin{table}[h]
\caption{Experimental results for various representative offline CIL methods in 6 tasks scenario (starting from learning the base task). Bold indicates the best performance in each metric.}
\centering
\smallskip\noindent
\resizebox{\linewidth}{!}{
\begin{tabular}{cc|cccc}
\toprule
\multicolumn{2}{c|}{\multirow{2}{*}{Method}} & \multicolumn{4}{c|}{CIFAR-100 w/ ResNet-32}\\
& & $A_f(\uparrow)$ & $A_a(\uparrow)$ & $F(\downarrow)$ & $A_l(\uparrow)$ \\
\midrule
\multirow{3}{*}{FT} & +BN & 35.86 & 45.71 & \textbf{37.14} & 77.32  \\
& +CN & 36.07 & 46.18 & 38.75 &{79.25}  \\ 
& +TBBN & \textbf{37.36} & \textbf{46.95} & {37.33} & \textbf{79.30} \\
\midrule
\multirow{3}{*}{EEIL} & +BN & 36.93 & 46.63 & 35.39 & 78.17 \\
& +CN & 38.44 & 47.45 & \textbf{33.91} & 78.23  \\ 
& +TBBN & \textbf{38.83} & \textbf{47.82} & {34.61} & \textbf{78.62}  \\ 
\midrule
\multirow{3}{*}{LUCIR} & +BN & 38.22 & 50.65 & \textbf{22.87} & 66.33  \\ 
& +CN & 38.20 & 49.74 & 24.99 & \textbf{68.32}  \\ 
& +TBBN & \textbf{39.54} & \textbf{50.95} & {23.27} & 67.57  \\
\midrule
\multirow{3}{*}{SSIL} & +BN & 45.69 & 53.03 & 8.55 & 53.63  \\ 
& +CN & 45.12 & 52.59 & \textbf{7.63} & 51.55  \\ 
& +TBBN &\textbf{46.61} & \textbf{53.48} & 8.94 & \textbf{55.16}  \\
\bottomrule
\end{tabular}
}
\vspace{-.2in}
\label{table:cil_baselines_base}
\end{table}


We conducted experiments for another CIL scenario, which involves learning half of all classes as the first task (base task) and then incrementally learning the remaining tasks, as proposed in \cite{(podnet)douillard2020podnet, (lucir)hou2019learning}.
We verified the effectiveness of TBBN for FT, LUCIR, and SS-IL on CIFAR-100 (with three seeds) in Table \ref{table:cil_baselines_base}. The scenario considered here involves six tasks, where the model learns 50 classes as the base task and then continues to learn five tasks, each consisting of 10 classes.
Note that this CIL scenario does not exactly correspond to the situation considered by TBBN (class-balanced tasks, see Section 3 of the manuscript). Nonetheless, the experimental results presented in Table \ref{table:cil_baselines_base} demonstrate that our TBBN can be successfully applied to several baselines, improving their performance compared to CN.

\subsection{Additional results from a 20-task setting}

\begin{table}[h]
\vspace{-.15in}
\caption{Experimental results (20 tasks).}
\vspace{-.15in}
\centering
\footnotesize
\smallskip\noindent
\resizebox{.8\linewidth}{!}{
\begin{tabular}{|cl||c|c|}
\hline
\multicolumn{2}{|c||}{$A_a$($\uparrow$)} & CIFAR-100 & ImageNet-100 \\ \hline \hline
\multicolumn{1}{|c|}{{}} & +BN & 29.66 & 39.06\\ 
\multicolumn{1}{|c|}{FT} & +CN($G=16$) & 30.12 & 37.82\\ 
\multicolumn{1}{|c|}{} & +TBBN & \textbf{34.45} & \textbf{43.84}\\
\hline
\multicolumn{1}{|c|}{{}} & +BN & 35.11 & 37.89\\ 
\multicolumn{1}{|c|}{EEIL} & +CN($G=16$) & 35.49 & 38.07\\ 
\multicolumn{1}{|c|}{} & +TBBN & \textbf{39.32} & \textbf{42.30}\\ 
\hline
\multicolumn{1}{|c|}{{}} & +BN & 34.36 & 39.34\\  
\multicolumn{1}{|c|}{LUCIR} & +CN($G=16$) & 34.83 & 36.54\\ 
\multicolumn{1}{|c|}{} & +TBBN & \textbf{37.07} & \textbf{39.90}\\ 
\hline
\multicolumn{1}{|c|}{{}} & +BN & 36.31 & 43.84\\  
\multicolumn{1}{|c|}{SSIL} & +CN($G=16$) & 36.00 & 43.12\\ 
\multicolumn{1}{|c|}{} & +TBBN & \textbf{38.55} & \textbf{46.08}\\ 
\hline
\end{tabular}
}
\vspace{-.15in}
\label{table:longer_tasks}
\end{table}

We also present additional results from a 20-task setting in Table \ref{table:longer_tasks}. 
We see that TBBN brings significant performance boost in various CIL scenarios. 

\section{Details of Experimental Settings}\label{section:sec4}
In the experiments using FT, EEIL~\cite{(eeil)castro2018end}, LUCIR~\cite{(lucir)hou2019learning}, and SSIL~\cite{(ss-il)ahn2021ss}, we followed the CIL benchmark code proposed by \cite{(cil_survey)masana2020class}. The network was trained using SGD with an initial learning rate of $10^{-1}$ and weight decay of $10^{-4}$, and a mini-batch size of 64. The number of epochs and schedule for adjusting the learning rate were set differently for each dataset and scenario. We used random sampling for ImageNet-100 experiments and herding~\cite{welling2009herding, (icarl)rebuffi2017icarl} for CIFAR-100 experiments. Table \ref{table:experimental_settings} provides detailed information on experimental settings and hardware used.
\begin{table}[h]
\caption{Details of experimental settings.}
\centering
\smallskip\noindent
\resizebox{.98\linewidth}{!}{
\begin{tabular}{|c||cc|cc|}
\hline
\multirow{2}{*}{}                                                                                                                                                                                & \multicolumn{2}{c|}{10 classes $\times$ 10 tasks}                                                                                                                                                                                                                                                  & \multicolumn{2}{c|}{5 classes $\times$ 20 tasks}                                                                                                                                                                                                                                    \\ \cline{2-5} 
                                                                                                                                                                                                 & \multicolumn{1}{c|}{CIFAR-100}                                                                                                                         & ImageNet-100                                                                                                                              & \multicolumn{1}{c|}{CIFAR-100}                                                                                                                       & ImageNet-100                                                                                                                 \\ \hline \hline
\begin{tabular}[c]{@{}c@{}}epochs per task\\ epoch for lr scheduling\\ lr decay\\ mini-batch size\\ model\\ python version\\ pytorch version\\ CUDA version\\ CuDNN version\\ GPU\end{tabular} & \multicolumn{1}{c|}{\begin{tabular}[c]{@{}c@{}}160\\ {[}80, 120{]}\\ 1/10\\ 64\\ ResNet-32\\ 3.7\\ 1.7.1+cu110\\ 11.2\\ 8.1.1\\ TITAN XP\end{tabular}} & \begin{tabular}[c]{@{}c@{}}100\\ {[}40, 80{]}\\ 1/10\\ 64\\ ResNet-18\\ 3.7\\ 1.7.1+cu110\\ 11.2\\ 8.1.1\\ RTX A5000\end{tabular} & \multicolumn{1}{c|}{\begin{tabular}[c]{@{}c@{}}160\\ {[}80, 120{]}\\ 1/10\\ 64\\ ResNet-32\\ 3.7\\ 1.7.1+cu110\\ 11.2\\ 8.1.1\\ 1080Ti\end{tabular}} & \begin{tabular}[c]{@{}c@{}}100\\ {[}40, 80{]}\\ 1/10\\ 64\\ ResNet-18\\ 3.7\\ 1.7.1+cu110\\ 11.2\\ 8.1.1\\ 1080Ti\end{tabular} \\ \hline
\end{tabular}
}
\label{table:experimental_settings}
\end{table}

In the case of experiments using PODNet~\cite{(podnet)douillard2020podnet} and AFC~\cite{(AFC)kang2022class}, we obtain the experimental results by implementing their official code. Also, we run each method with the default hyperparameter setting proposed in the official code.

\section{Pseudo code of TBBN}\label{section:sec1.1}
Algorithm \ref{alg:tbbn} shows the Pytorch-style pseudo algorithm for TBBN's forward function. It is important to note that TBBN does not require hyperparameter tuning and only uses easily accessible information such as the task number ($t$) and the number of sampled current ($B_c$) and memory ($B_p$) data.


\begin{algorithm*}[h]
        \caption{Pytorch-style pseudo algorithm of the forward function of TBBN. Note that running\_mean and running\_val ($\in\textbf{R}^C$) are initialized to $0$ and $1$, and gamma and beta ($\in\textbf{R}^C$) are initialized to $1$ and $0$, respectively.  m is the hyperparameter for the exponential moving average of running mean and standard deviation, and we set it to 0.9}\label{alg:tbbn}
        \begin{algorithmic}[1]
            \STATE def \textbf{TBBN\_forward}(x$,$ t$, B_c, B_p,$ train):
            
            \STATE \hskip1.0em if train:
            \STATE \hskip2.em if t $\neq 1$: \ \ \ \textcolor{gray}{\# set r} 
            \STATE \hskip3.0em r $ = B_c / B_p \cdot (t-1)$
            \STATE \hskip2.em else:
            \STATE \hskip3.0em r $ = 1$
            \STATE \hskip2em if r not in CD($B_c, B_p$):  \ \ \ \textcolor{gray}{\# find r$^*$, CD returns common divisor of given two values (check Equation (1))}
            \STATE \hskip3.0em r $ = M(B_c, B_p, $ r$)$
            \STATE \hskip2.em 
            \STATE \hskip2.em $N, C, H, W = $x.shape 
            \STATE \hskip2.em \textcolor{gray}{\# make balanced batch}
            \STATE \hskip2.em curr$\_$batch = x[:$B_c$].reshape($B_{c}/$r$, C\cdot $r$, H, W$)
            \STATE \hskip2.em prev$\_$batch = x[$B_c$:].repeat($1, $r$, 1, 1$)
            \STATE \hskip2.em bal$\_$batch = concat((curr$\_$batch, prev$\_$batch), dim $= 1$)
            \STATE \hskip2.em 
            \STATE \hskip2.em \textcolor{gray}{\# calculate balanced mean and variance}
            \STATE \hskip2.em bal$\_$mean = bal$\_$batch.mean(dim $= [0,2,3]$)
            \STATE \hskip2.em bal$\_$val = bal$\_$batch.val(dim $= [0,2,3]$)
            \STATE \hskip2.em
            \STATE \hskip2.em \textcolor{gray}{\# normalize reshaped input batch}
            \STATE \hskip2.em bal$\_$batch$ = ($bal$\_$batch $ - $ bal$\_$mean$) / ($bal$\_$val $ + \epsilon)$.sqrt()
            \STATE \hskip2.em \textcolor{gray}{\# affine-transform the normalized batch}
            \STATE \hskip2.em bal$\_$batch$ = $bal$\_$batch$*$gamma.repeat(r) $ + $ beta.repeat(r)
            \STATE \hskip2.em 
            \STATE \hskip2.em \textcolor{gray}{\# reshape bal$\_$batch to the original shape}
            \STATE \hskip2.em bal$\_$batch$\_$curr = bal$\_$batch[$:B_c/$r].reshape($B_{c}, C, H, W$)
            \STATE \hskip2.em bal$\_$batch$\_$prev = bal$\_$batch[$B_c/$r$:$].reshape($B_{p}, $ r$, C , H, W$).mean(dim=1)
            \STATE \hskip2.em x = concat((bal$\_$batch$\_$curr, bal$\_$batch$\_$prev), dim=1)
            \STATE \hskip2.em 
            \STATE \hskip2.em \textcolor{gray}{\# update running\_mean and running\_val}
            \STATE \hskip2.em running$\_$mean$\_$temp $ = $ running$\_$mean.repeat(r)
            \STATE \hskip2.em running$\_$val$\_$temp $ = $ running$\_$val.repeat(r)
            \STATE \hskip2.em running$\_$mean$\_$temp $ = $ m $ * $ running$\_$mean$\_$temp $  +  (1-$m$) * $ bal$\_$mean
            \STATE \hskip2.em running$\_$val$\_$temp $ = $ m $ * $ running$\_$val$\_$temp $  +  (1-$m$) * $ bal$\_$val
            \STATE \hskip2.em 
            \STATE \hskip2.em running$\_$mean $ = $ running$\_$mean$\_$temp.reshape(r, $C$).mean(dim = 0)
            \STATE \hskip2.em running$\_$val $ = $ running$\_$val$\_$temp.reshape(r, $C$).mean(dim = 0)
            \STATE \hskip2.em 
            
            \STATE \hskip1.em 
            \STATE \hskip1.0em else:
            \STATE \hskip2.em x$ = ($x$ - $ running$\_$mean$) / ($running$\_$val $+\epsilon)$.sqrt()
            \STATE \hskip2.em x$ = $x$*$gamma $ + $ beta
            \STATE \hskip2.em 
            \STATE \hskip1.0em return x
            
        \end{algorithmic}
\end{algorithm*}

\section{Experiments for Online CL.} 
Table \ref{table:onlinecl} presents the online CL results for CIFAR-100. We follow the experimental settings (ResNet-18, 20 tasks, single epoch and 2000 exemplars) proposed in \cite{(online_cl_survey)mai2021online}, and only conduct experiments using finetuning (FT) for comparison. 
\begin{table}[h]
\vspace{-.1in}
\caption{Experimental results (20 tasks).}
\vspace{-.1in}
\centering
\footnotesize
\smallskip\noindent
\resizebox{.8\linewidth}{!}{
\begin{tabular}{|c|l||c|c|}
\hline
\multicolumn{2}{|c||}{$A_a$($\uparrow$)} & Class-IL & Task-IL \\ 
\hline \hline
 \multirow{4}*{FT} & +BN & 10.77 & 64.39 \\ 
 & +CN ($G=8$) & \textbf{10.94} & \textbf{68.70} \\ 
 & +CN ($G=16$) & 8.43 & 64.23 \\ 
 & +TBBN & 10.12 & 67.43 \\ 
 \hline
\end{tabular}
}
\vspace{-.1in}
\label{table:onlinecl}
\end{table}
Our results show that CN with 
($G=8$) outperforms TBBN in both class-IL and task-IL. However, the performance gain of CN for class-IL is not as substantial as in \cite{(CN)pham2022continual}
This trend was also shown previously with SplitTinyIMN in Table 4 of \cite{(CN)pham2022continual}.

\newpage

{\small
\bibliographystyle{plain}
\bibliography{example_paper.bib}
}